\documentclass[pdflatex,sn-basic]{sn-jnl}% Basic Springer Nature Reference Style/Chemistry Reference Style

\jyear{2021}%

\theoremstyle{thmstyleone}%
\theoremstyle{thmstyletwo}%

\theoremstyle{thmstylethree}%

\usepackage{tabularx}
\usepackage{rotating}
\usepackage{multirow}
\usepackage{multicol}
\usepackage{makecell}
\usepackage{graphicx}
\usepackage{enumitem}
\usepackage[normalem]{ulem}
\usepackage[font=footnotesize]{subfig}
\usepackage{amsthm}
\theoremstyle{plain}
\newtheorem{thm}{Theorem}% reset theorem numbering for each chapter
\theoremstyle{definition}
\newtheorem{defn}[thm]{Definition} % definition numbers are dependent on theorem numbers
\usepackage[pagewise]{lineno}
\usepackage[symbol]{footmisc}   % library for footnotes

\begin{document}

\title[GCN-M: GCNs for Traffic Forecasting with Missing Values]{Graph Convolutional Networks for Traffic Forecasting with Missing Values}
\artnote{This research has been developed for the most part in the context of the main author’s Ph.D. at DAVID Lab, UVSQ, Université Paris-Saclay.}

\author*[1]{\fnm{Jingwei} \sur{Zuo}}\email{jingwei.zuo@tii.ae}
\author[2]{\fnm{Karine} \sur{Zeitouni}}\email{karine.zeitouni@uvsq.fr}

\author[2]{\fnm{Yehia} \sur{Taher}}\email{yehia.taher@uvsq.fr}

\author[3]{\fnm{Sandra} \sur{Garcia-Rodriguez}}\email{sandra.garciarodriguez@cea.fr}

\affil[1]{\orgname{Technology Innovation Institute}, \orgaddress{\city{Abu Dhabi}, \country{UAE}}}
\affil[2]{\orgdiv{DAVID Lab, UVSQ}, \orgname{Université Paris-Saclay}, \orgaddress{\city{Versailles}, \country{France}}}
\affil[3]{\orgdiv{Data Analysis and Systems Intelligence Laboratory}, \orgname{CEA, LIST}, \orgaddress{\city{Gif Sur Yvette}, \country{France}}}


\abstract{Traffic forecasting has attracted widespread attention recently. In reality, traffic data usually contains missing values due to sensor or communication errors. The Spatio-temporal feature in traffic data brings more challenges for processing such missing values, for which the classic techniques (e.g., data imputations) are limited: 1) in temporal axis, the values can be randomly or consecutively missing; 2) in spatial axis, the missing values can happen on one single sensor or on multiple sensors simultaneously.
%Different from previous work which ignore the missing values during the training process, we claim and demonstrate that the Spatio-temporal feature in traffic data brings more challenges: the values can be randomly or consecutively missing, on one single sensor or on multiple sensors, which limit the classic techniques for processing missing values, such as data imputation.
%2, Previous work & motivation
Recent models powered by Graph Neural Networks achieved satisfying performance on traffic forecasting tasks. However, few of them are applicable to such a complex missing-value context.
%5, Our test and result
To this end, we propose GCN-M, a Graph Convolutional Network model with the ability to handle the complex missing values in the Spatio-temporal context. Particularly, we jointly model the missing value processing and traffic forecasting tasks, considering both local Spatio-temporal features and global historical patterns in an attention-based memory network. We propose as well a dynamic graph learning module based on the learned local-global features.
The experimental results on real-life datasets show the reliability of our proposed method.
%The experimental results on real-life datasets show the reliability of our model.
}

\keywords{Traffic Forecasting, Missing Values, Graph Convolutional Networks, Memory Networks, Neural Networks, Deep Learning} 



\maketitle
\section{Introduction}
Traffic forecasting has played a critical role in intelligent transportation systems, which helps the transportation department better manage and control traffic congestion. 
Generally represented by geo-located Multivariate Time Series (MTS), traffic data not only shows the typical characteristics of MTS, i.e., temporal dependency \citep{zuo2021smate}, but also integrates the spatial information of the traffic network, i.e., the spatial dependency between the sensor traffic nodes over the road network. 

In recent years, by leveraging the spatial-temporal patterns in traffic data, many deep learning models based on recurrent neural network (RNN) \citep{li2018diffusion}, temporal convolutional network (TCN) \citep{wu2019graph}, graph convolutional networks (GCN) \citep{li2021dynamic}, etc., have been applied in traffic forecasting tasks and achieved state-of-the-art performance.
They all have a strong assumption that the data is complete or has been well-preprocessed \citep{yu2018spatio}. However, since the traffic data is generally collected from geo-located sensors, sensor failures or communication errors will result in missing values in the collected data, thus deteriorating the performance of the forecasting model. 
We should remark that the missing measures are usually marked as \textit{zero} in traffic data \citep{li2021dynamic}, which should be distinguished from the non-missing measures but with \textit{zero} values. A typical example \citep{tian2018lstm} comes from the traffic flow data: no vehicles are detected during the night, then the traffic measures are marked as \textit{zero} instead of being considered as missing. \textcolor{black}{This can be commonly observed from real-life traffic data for which the missing rate evolves periodically during the day \citep{lopez2018traffic}.} 

\begin{figure}[htbp]
\centering
\includegraphics[width=300pt]{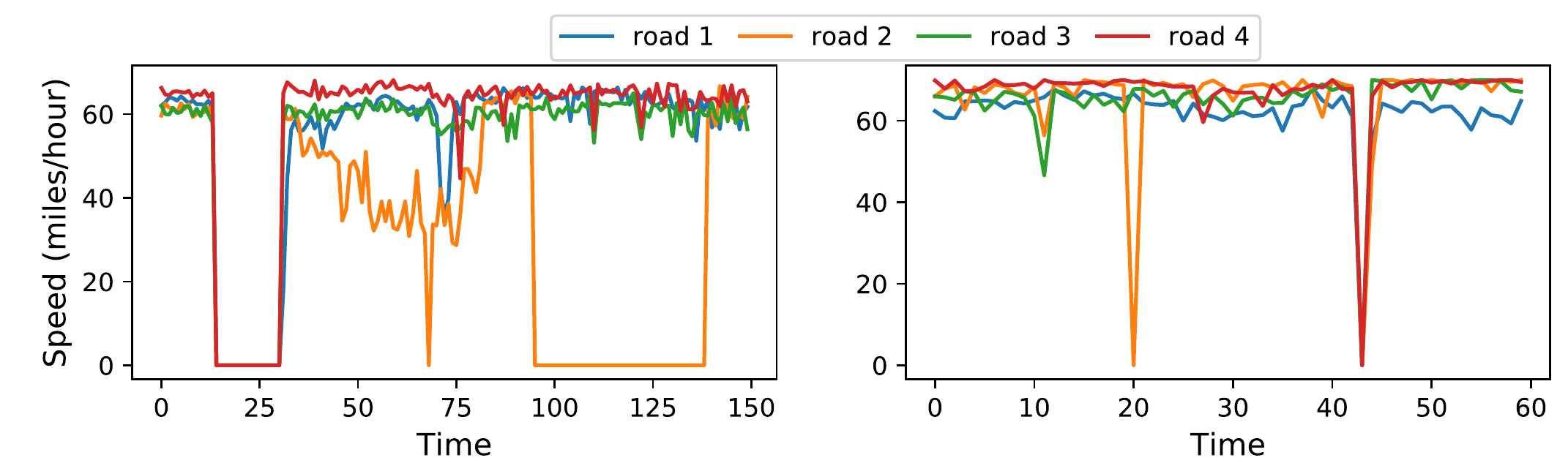}
\caption{Missing measures of traffic speed data from METR-LA dataset \citep{li2018diffusion}. \textit{left}) long-range missing on entire network (i.e., Spatio-temporal block) and partial network (i.e., temporal block); \textit{right}) short-range random missing on partial network (i.e., temporal values) and entire network (i.e., Spatio-temporal vectors).}
\vspace{-1em}
\label{MissingValueScenarios}
\end{figure}

The missing values can either be ignored in the learning model when calculating the loss function \citep{wang2020traffic} or be considered before or during the training process \citep{cui2020graph}. Ignoring the missing values, especially when the missing ratio is high \citep{cui2020graph}, hinders the model from benefiting from the rich data information for better performance. When considering the missing values in traffic data, most work \citep{cirstea2019graph} conducts data imputation during the preprocessing step, then imports the completed data into the training step, i.e., \textit{two-step processing}. Recent work tends to jointly consider the missing values and the forecasting modeling during the training step (i.e., \textit{one-step processing}) and declared better performance than the two-step processing \citep{che2018recurrent,cui2020stacked,cui2020graph,tian2018lstm,tang2020joint}.
However, the above-mentioned work suffers from three major issues. First, the missing and zero values are usually considered to be the same, leading to unnecessary, even harmful data imputations, thus contradicting the raw data information. Second, most of the work \citep{che2018recurrent,cui2020stacked,tian2018lstm,tang2020joint} considers missing values from the temporal aspect, ignoring the rich information from the spatial perspective. Third, they are generally designed for processing the missing values in some basic scenarios, such as random missing values or temporal block missing values, but lack power for the complex scenarios as shown in Fig. \ref{MissingValueScenarios}. In the real world, the missing values in traffic data occur in both long-range (e.g., \textit{device power-off}) and short-range (e.g. \textit{device errors}) settings, in partial (e.g., \textit{local sensor errors}) and entire transportation network (e.g., \textit{control center errors}). Therefore, a holistic approach is required for handling various types of missing values together in complex scenarios.

To handle both the Spatio-temporal patterns and complex missing-value scenarios in traffic data, we propose \textbf{G}raph \textbf{C}onvolutional \textbf{N}etworks for Traffic Forecasting with \textbf{M}issing Values (GCN-M). The graph neural network-based structure allows jointly modeling the Spatio-temporal patterns and the missing values in a one-step process. We construct local statistical features from spatial and temporal perspectives for handling short-range missing values. This is further enhanced by a memory module to extract global historical features for processing long-range missing blocks. The combined local-global features allow not only for identifying the missing measures from the inherent zero values but also for enriching the traffic embeddings, thus generating dynamic traffic graphs to model the dynamic spatial interactions between traffic nodes. The missing values on a partial and entire network can then be considered from spatial and temporal perspectives.

We summarize the paper's main contributions as follows:
\begin{itemize}
    \item[$\bullet$]\textbf{Complex missing value modeling}: We study the complex scenario where missing traffic values occur on both short \& long ranges and on partial \& entire transportation networks. 
    \item[$\bullet$]\textbf{Spatio-temporal memory module}: We propose a memory module that can be used by GCN-M to learn both local Spatio-temporal features and global historical patterns in traffic data for handling the complex missing values. 
    \item[$\bullet$]\textbf{Dynamic graph modeling}: We propose a dynamic graph convolution module that models the dynamic spatial interactions. The dynamic graph is characterized by the learned local-global features at each timestamp, which not only offset the missing values' impact but also help learn the graph.
    \item[$\bullet$]\textbf{Joint model optimization}: We jointly model the Spatio-temporal patterns and missing values in one-step processing, which allows processing missing values specifically for traffic forecasting tasks, thus bringing better model performance than two-step processing.
    \item[$\bullet$]\textbf{Extensive experiments on real-life data}: The experiments are carried out on two real-life traffic datasets. We provide detailed evaluations with 12 baselines, which show the effectiveness of GCN-M over state-of-the-art.
\end{itemize}
The rest of this paper starts with a review of the most related work. Then, we formulate the problems of the paper. Later, we present in detail our proposal GCN-M, followed by the experiments on real-life datasets and the conclusion.

\section{Related Works}
\color{black}
We start with defining the notions used in the paper:

\begin{defn}(One-step processing). For one-step processing models, the missing values and the traffic forecasting are jointly modeled in one single step. 
\end{defn}
\begin{defn}(Two-step processing). The two-step processing models first handle the missing values in a preprocessing step, then apply a forecasting model on the completed data.
\end{defn}
\color{black}
\subsection{Graph Convolutional Networks for Traffic Forecasting}
Graph Convolutional Network (GCN) is a special kind of Convolutional Neural Network (CNN) generalized for graph-structured data. Most of the GCN-related work focuses on graph representation, which learns node embedding by integrating the features from the node's local neighbors based on the given graph structure, i.e., adjacency matrix. The traffic data shows strong dependencies between the spatial nodes, for which GCN can be naturally suitable. Various work \citep{li2018diffusion,wu2020connecting,yu2018spatio,wang2020traffic} empowered by GCN achieved remarkable performance when doing traffic forecasting tasks, relying on spatial and temporal completion of the data, or calculating loss function for non-zero entries, i.e., only calculating the loss on entries that contain valid sensor readings. However, these techniques may introduce derivations when modeling the Spatio-temporal relations between the sensor nodes. In other words, where non-missing measures are required to characterize the dynamic graph at each timestamp, missing values may hinder the traffic graph learning \citep{li2021dynamic}, especially dynamic graph learning \citep{guo2021learning}.

\subsection{Missing value processing}
The simplest solution for processing missing values in MTS would be data imputation, such as statistic imputation (e.g., mean, median), EM-based imputation \citep{garcia2010pattern}, K-nearest neighborhood \citep{batista2002study}, and matrix factorization \citep{dong2022laplacian}. It's generally believed that those methods fail to model temporal dynamics in a time series \citep{tang2020joint}. In other words, they are not applicable for handling long-range missing values. Recent generative models \citep{yoon2019time,dong2022laplacian} show reliable performance for long-range time series imputation. However, isolating the imputation model from the forecasting model leads to two-step processing, and may generate sub-optimal results \citep{cirstea2019graph,wells2013strategies,che2018recurrent}. To handle this issue, recent studies \citep{che2018recurrent,tang2020joint,wang2021fine,zhong2021heterogeneous} jointly model the missing values and forecasting task in one-step processing. For instance, GRU-D \citep{che2018recurrent} considers the nearby temporal statistical features to do imputations inside GRUs, whereas LSTM-I \citep{cui2020stacked} infers missing values at the current time step from preceding LSTM cell states and hidden states, and SGMN \citep{cui2020graph} improved the state transition process via a Graph Markov Process. Limited to short-period missing context, those methods are further enhanced by LGnet \citep{tang2020joint} with the global temporal dynamics to handle the long-range missing issue, and by LSTM-M \citep{tian2018lstm} with multi-scale modeling to better explore historical information. However, the above-mentioned models handle missing values by focusing on the temporal aspect without considering the complex Spatio-temporal features in traffic data. Specifically, the strong spatial connections between the sensor nodes should provide us with more information to handle the missing values. Moreover, one-step processing models are generally designed for single-step forecasting without considering the multi-step settings. Table \ref{method_comparison} shows the method comparison for traffic forecasting with missing values.

\begin{table}[htbp]
\centering
\caption{Existing methods for Traffic Forecasting with missing values}
\label{method_comparison}
\scalebox{0.8}{
\begin{tabular}{cccccccc} 
\toprule
                       & Imputation-based & GRU-D        & LSTM-I       & LSTM-M       & LGnet        & SGMN         & \textbf{GCN-M}  \\ 
\midrule
Short-range missing    & $\checkmark$     & $\checkmark$ & $\checkmark$ & $\checkmark$ & $\checkmark$ & $\checkmark$ & $\checkmark$    \\
Long-range missing     & -                & -            & -            & $\checkmark$ & $\checkmark$ & -            & $\checkmark$    \\
Multi-scale modeling   & -                & -            & -            & $\checkmark$ & -            & -            & $\checkmark$    \\
Spatial modeling       & -                & -            & -            & -            & -            & $\checkmark$ & $\checkmark$    \\
One-step processing    & -                & $\checkmark$ & $\checkmark$ & $\checkmark$ & $\checkmark$ & $\checkmark$ & $\checkmark$    \\
Multi-step forecasting & $\checkmark$     & -             & -             & -             & $\checkmark$ & -              & $\checkmark$    \\
\bottomrule

\end{tabular}
}
\vspace{-1em}
\end{table}

\section{Problem Formulation}
\label{Problem Formulation}
We aim to predict future traffic data by leveraging historical traffic data. Traffic data can be represented as a multivariate time series on a traffic network. Let the traffic network $\mathcal{G=\{V, E\}}$, where $\mathcal{V}=\{v_1, ..., v_N\}$ is a set of $N$ traffic sensor nodes and $\mathcal{E}=\{e_1, ..., e_E\}$ is a set of $E$ edges connecting the nodes. Each node contains $F$ features representing traffic flow, speed, occupancy, etc.
We use $\mathcal{X}$=$\{\mathbf{X}_t\}_{t=1}^{\tau}\in \mathcal{R}^{N \times F \times \tau }$ to denote all the feature values of all the nodes over $\tau$ time slices, $\mathbf{X}_{t}=(\textbf{x}_{t}^{1},...,\textbf{x}_{t}^{N}) \in \mathcal{R}^{N \times F}$ denotes the observations at time $t$, where $\textbf{x}_t^{i}\in \mathcal{R}^{F}$ is the $i$-th variable of $X_t$.
We define a mask sequence $\mathcal{M}$=$\{\mathbf{M}_t\}_{t=1}^{\tau}\in \mathcal{R}^{N \times F \times \tau }$, $\mathbf{M}_{t}=(\textbf{m}_{t}^{1},...,\textbf{m}_{t}^{N}) \in \mathcal{R}^{N \times F}$. $\textbf{m}_t^{i} \in \{0,1\}^{F}$ denotes the features' missing status for the $i$-th variable.
To simplify, we adopt $x_t^i \in \mathcal{R}$ and $m_t^i \in \mathcal{R}$ to denote respectively the observation and mask value of \textit{one single feature} for the $i$-th variable of $\mathbf{X}_t$. We take $m_t^i=0$ if $x_t^i$ is missing, otherwise $m_t^i=1$.

We aim to build a model $f$, which can take an incomplete traffic sequence \{$\mathcal{X}$, $\mathcal{M}$\} and the traffic network $\mathcal{G}$ as input, to predict the traffic data for the next $T_{p}$ time steps $\mathbf{Y}=\{y_{\tau+1}, ..., y_{\tau+T_{p}}\} \in \mathcal{R}^{N\times T_{p}}$.

\section{Proposal: GCN-M}
Traffic data is collected under complex urban conditions. Apart from the Spatio-temporal patterns in the traffic data, we also consider the scenarios of complex missing values. We design a solution that models the local Spatio-temporal features and global historical patterns in a dynamic manner. The complex missing values are considered when building the forecasting model, i.e., one-step processing.

\subsection{Model Architecture}

\begin{figure}[htbp]
\centering
\includegraphics[width=350pt]{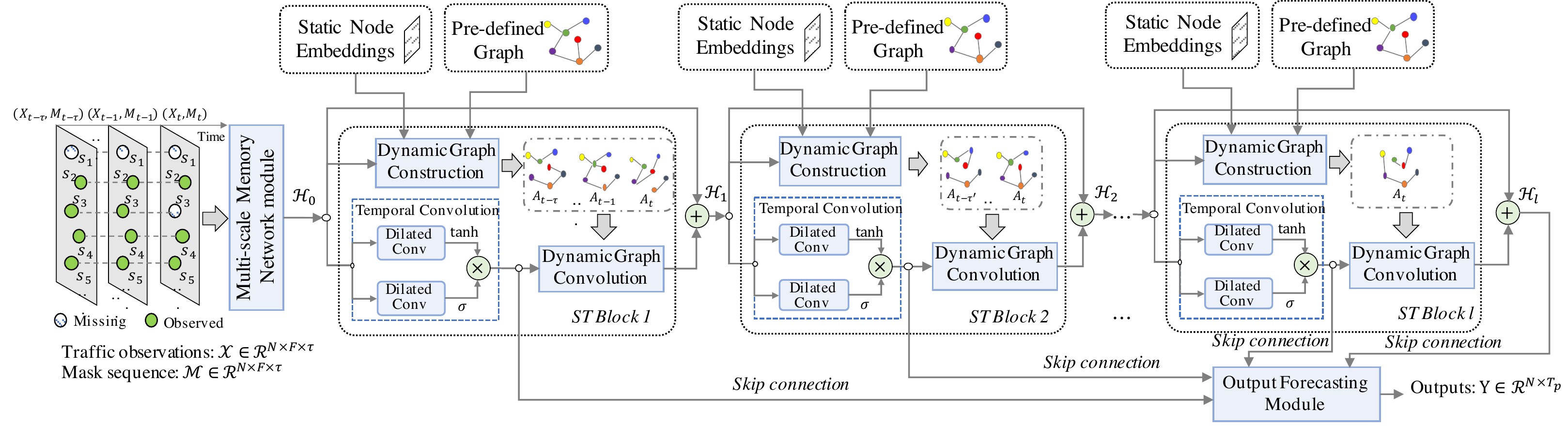}
\caption{Main architecture of GCN-M}
\label{SystemStructure}
\end{figure}

The global structure of GCN-M is shown in Fig. \ref{SystemStructure}, integrating a Multi-scale Memory Network module, an Output Forecasting module, and $\textit{l}$ Spatio-Temporal (ST) blocks. Each ST block integrates three key components: Temporal Convolution, Dynamic Graph Construction, and Dynamic Graph Convolution. 
The input traffic observations $\mathcal{X}\in \mathcal{R}^{N \times F \times \tau }$ and the mask sequence $\mathcal{M}\in \mathcal{R}^{N \times F \times \tau }$ are fed into the multi-scale memory network to extract the local statistic features and global historical patterns thus enriching the traffic embeddings. 
On the one hand, the enriched embeddings $\mathcal{H}_i$ on each ST block are used to mark the dynamic traffic status, thus generating dynamic graphs by combining both static node embeddings and predefined graph information. On the other hand, the learned dynamic graphs are combined with the temporal convolution module via a dynamic graph convolution to capture temporal and spatial dependencies in the traffic embeddings.
We adopt residual connections between the input and output of each ST block to avoid the gradient vanishing problem. 
The output forecasting module takes the skip connections on the output of the final ST block and the hidden states after each temporal convolution for final prediction.

\subsection{Multi-scale Memory Network}
\begin{figure}[htbp]
\centering
\includegraphics[width=360pt]{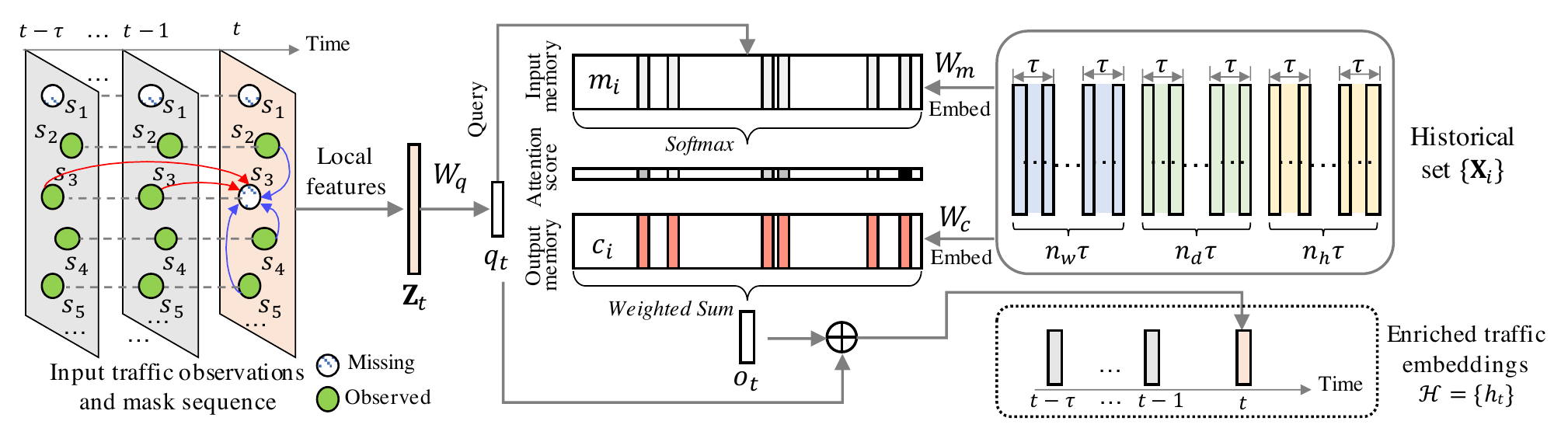}
\caption{Memory module enriches traffic embeddings with multi-scale global features}
\label{MemoryModule}
\end{figure}

%General idea of Memory network
To extract the local statistic features and global historical patterns then form an enriched embedding, we adopt the concept of memory network, which was firstly proposed in \citep{weston2015memory} with primary application in Question-Answer (QA) systems. As shown in Fig. \ref{MemoryModule}, the main idea of our memory network is to learn from historical memory components which conserve the long-range multi-scale patterns, i.e., recent, daily-periodic, and weekly-periodic dependencies. The scale range depends on the data characteristics.
Specifically, we first extract local Spatio-temporal features as keys to query the memory components; the weighted historical long-range patterns will be cooperated with the local statistic features to eliminate the side effect from the missing values. Then, the local-global features will be output as the enriched traffic embeddings. 

\subsubsection{Local Spatio-temporal features}
We first extract the Spatio-temporal features using the contextual information from observed parts of the time series. Unlike prior studies \citep{che2018recurrent}, we consider both temporal and spatial aspects for generating the following statistic features of every timestamp:

\textit{Empirical Temporal Mean}: The mean of previous observations reflects the recent traffic state and serves as a contextual knowledge of $x_{t}^{i}$. Therefore, for a missing value $x_{t}^{i}\in \mathcal{R}$, we construct its temporal mean using $L$ \textcolor{black}{past} samples $x_*^{i}$ before time $t$:
\begin{equation}
\label{temp_mean_equation}
\small
    \bar{x}_t^i=\sum_{l=t-L}^{t-1}m_{l}^{i}x_{l}^{i}/\sum_{l=t-L}^{t-1}m_{l}^{i}
\end{equation} 

\textit{Last Temporal Observation}: We adopt the assumption in \citep{che2018recurrent} that any missing value inherits more or less the information from the last non-missing observation. In other words, the temporal neighbor stays close to the current missing value. We use $\dot{x}_{t}^{i}$ to  denote the last temporal observation of $x_{t}^{i}$, their temporal distance is defined as $\dot{\delta}_{t}^{i}$.

\textit{Empirical Spatial Mean}: Another contextual knowledge of $x_{t}^{i}$ is from the nearby nodes, which reflects the current local traffic situation. For each missing value $x_t^i$, we construct its empirical spatial mean using $S$ \textcolor{black}{nearby} samples $x_t^*$ of the sensor node $i$:
\begin{equation}
\label{spatial_mean_equation}
\small
    \bar{\bar{x}}_{t}^{i}=\sum_{s=1}^{S}m_{t}^{s}x_{t}^{s}/\sum_{s=1}^{S}m_{i}^{s}
\end{equation}

\textit{Nearest Spatial Observation}: Typically, the state of a graph node remains relatively similar to its neighbors, especially in a traffic graph where the nearby nodes share similar traffic situations. We define $\ddot{x}_{t}^{i}$ as the nearest spatial observation of $x_{t}^{i}$, their spatial distance is denoted as $\ddot{\delta}_{t}^{i}$.

Generally, when $\dot{\delta}_{t}^{i}$ or $\ddot{\delta}_{t}^{i}$ is smaller, we tend to trust $\dot{x}_{t}^{i}$ or $\ddot{x}_{t}^{i}$ more. When the spatial/temporal distance becomes larger, the spatial/temporal mean would be more representative. Under this assumption, we model the temporal and spatial decay rate $\gamma$ as 
\begin{equation}
\small 
    \gamma_{t}(\dot{\delta}_{t}^{i}) = exp\{-max(0, w^{i}\dot{\delta}_{t}^{i} + b^{i}\}
\end{equation}
\begin{equation}
\small
    \gamma_{s}(\ddot{\delta}_{t}^{i}) = exp\{-max(0, w_{t}\ddot{\delta}_{t}^{i} + b_{t}\}
\end{equation}
where $w^i$, $w_t$, $b^i$ and $b_t$ are model parameters that we train jointly with other parameters of the traffic forecasting model. We chose the exponentiated negative rectifier \citep{che2018recurrent} so that the decay rates $\gamma_{t}$ and $\gamma_{s}$ decrease monotonically in the range between 0 and 1. Considering the trainable decays, our proposed model incorporates the spatial/temporal estimations to define the local features of $x_{t}^{i}$:
\begin{equation}
\small
    z_{t}^{i}= m_{t}^{i}x_{t}^{i} + (1-m_{t}^{i})(
    \gamma_{t}\dot{x}_{t}^{i} +
    \gamma_{s}\ddot{x}_{t}^{i} +
    (1-\gamma_{t})\bar{x}_{t}^{i} + 
    (1-\gamma_{s})\bar{\bar{x}}_{t}^{i}
    )
\end{equation}
Therefore, for $\mathbf{X}_t\in \mathcal{R}^{N \times F}$, we can get its local features $\mathbf{Z}_t\in \mathcal{R}^{N \times F}$.

\subsubsection{Multi-scale Memory Construction}
Global historical patterns play a critical role in building an enriched traffic embedding. The historical observations in multiple scales (e.g., hourly, daily, weekly) can be embedded into memory as complement information for the local features $\mathbf{Z}_{t}\in \mathcal{R}^{N\times F}$. The main idea is to adopt local features to query similar historical patterns in the memory and output a weighted feature representation for the current timestamp. In this manner, the enriched multi-scale historical and local features allow not only eliminating the side effect of missing values but also improving the current feature embeddings.
At time $t$, the query $q_{t}$ of $\mathbf{X}_{t}$ can be embedded from the local features $\mathbf{Z}_{t}\in \mathcal{R}^{N\times F}$:
\begin{equation}
\small
    q_{t}= \mathbf{Z}_{t}W_{q} + b_{q} \in \mathcal{R}^{N \times d}
\end{equation}
where $W_{q}\in \mathcal{R}^{F\times d}$, $b_{q}\in \mathcal{R}^{N\times d}$ are parameters, $d$ is the embedding dimension.

The input memory components are the temporal segments of multiple scales:
\begin{itemize}
    \item[$\bullet$] The recent (e.g., hourly) segment is: $X_h$ = $\{\mathbf{X}_{i}\}_{i=t-\tau}^{t-1} \in \mathcal{R}^{N \times F \times n_{h}\tau}$,  with $n_{h}$ recent periods (e.g., hours) before $t$, each period contains $\tau$ observations.
    \item[$\bullet$] The daily-periodic segment is: $X_d$ = $\{\mathbf{X}_i\}\in \mathcal{R}^{N \times F \times n_{d}\tau}$ with $i\in $ $[t-n_{d}T_{d}-\tau/2:t-n_{d}T_{d}+\tau/2]$ $\|$ $[t-(n_{d}-1)T_{d}-\tau/2:t-(n_{d}-1)T_{d}+\tau/2]$ $\|$ ... $\|$ $[t-T_{d}-\tau/2:t-T_{d}+\tau/2]$, we store $\tau$ samples around time $t$ for each of the past $n_d$ days. $T_d$ denotes the sample number during one day, and $\|$ indicates the concatenation operation.
    %$X_d$ = $\{x_t\}$ with $t\in $ $[t-n_{d}T_{d}-n_{h}\tau:t-n_{d}T_{d}-1]$ $\|$ $[t-(n_{d}-1)T_{d}-n_{h}\tau:t-(n_{d}-1)T_{d}-1]$ $\|$ ... $\|$ $[t-T_{d}-n_{h}\tau:t-T_{d}-1]$, 
    \item[$\bullet$] The weekly-periodic segment is: $X_w$ = $\{\mathbf{X}_i\}\in \mathcal{R}^{N \times F \times n_{w}\tau}$ with $i\in $ $[t-n_{w}T_{w}-\tau/2:t-n_{w}T_{w}+\tau/2]$ $\|$ $[t-(n_{w}-1)T_{w}-\tau/2:t-(n_{w}-1)T_{w}+\tau/2]$ $\|$ ... $\|$ $[t-T_{w}-\tau/2:t-T_{w}+\tau/2]$, we store $\tau$ samples around time $t$ for each of the past $n_w$ weeks. $T_w$ denotes the sample number during one week, and $\|$ indicates the concatenation operation.
\end{itemize}
The input set of $\{\mathbf{X}_i\}$ = $[X_h \| X_d \| X_w] \in \mathcal{R}^{N \times F \times (n_d+n_w+n_h)\tau}$ are embedded into the input memory vectors $\{m_i\}$ and output memory vectors $\{c_i\}$:
\begin{equation}
\small
    m_{i}= \mathbf{X}_{i}W_{m} + b_{m}  \in \mathcal{R}^{N \times d}
\end{equation}
\begin{equation}
\small
    c_{i}= \mathbf{X}_{i}W_{c} + b_{c}  \in \mathcal{R}^{N \times d}
\end{equation}
where $W_{m},W_{c}\in \mathcal{R}^{F\times d}$, $b_{m},b_{c}\in \mathcal{R}^{N\times d}$ are parameters.

In the embedding space, we compute the attention score between the query $q_{t}$ and each memory $m_{i}$ by taking the inner product followed by a Softmax:
\begin{equation}
\small
    p_{t,i}= Softmax(q_{t}^{T}m_{i})
\end{equation}
The attention score represents the similarity of each historical observation to the query. Any pattern with a higher attention score is more similar to the context of targeting missing values. As shown in Fig. \ref{MemoryModule}, the response vector from memory is then a sum over the output memory vectors, weighted by the attention score from the input:
\begin{equation}
\small
    o_{t}= \textstyle\sum_{i=1}^{(n_d+n_w+n_h)\tau} c_{i} p_{t,i} \in \mathcal{R}^{N\times d}
\end{equation}

We can finally integrate both local Spatio-temporal and global multi-scale features and output the enriched traffic embeddings:
\begin{equation}
\small
    h_{t}= (q_{t} \| o_{t})W_{h} + b_{h} \in \mathcal{R}^{N\times d}
\end{equation}
where $W_{h}\in \mathcal{R}^{2d\times d}$, $b_{h}\in \mathcal{R}^{d}$ are parameters, and $\|$ denotes the concatenation operation. 
Therefore, for input $\mathcal{X}=\{\mathbf{X}_t\}_{t=1}^{\tau}\in \mathcal{R}^{N \times F \times \tau }$, we can get its enriched traffic embeddings $\mathcal{H}=\{h_t\}_{t=1}^{\tau}\in \mathcal{R}^{N \times d \times \tau }$.

\subsection{Dynamic Graph Construction}
\label{Dynamic Graph Construction}
A predefined graph is usually constructed with the distance or the connectivity between the spatial nodes. However, recent studies \citep{wang2020traffic,wu2020connecting,li2021dynamic} show that the cross-region dependence does exist for those nodes which are not physically connected but share similar patterns. Learning dynamic graphs should show better performance than learning static graphs or adopting the predefined graphs.
% add more motivations for doing so!
Considering the missing values in traffic data, instead of using the raw traffic observations to mark the dynamic traffic status \citep{li2021dynamic,han2021dynamic}, we construct dynamic graphs (i.e., adjacency matrix) with the enriched traffic embeddings $\mathcal{H}_i$ at each ST block, which integrates both local and global multi-scale patterns at each time step. This allows capturing the spatial relationship between traffic nodes robustly. As shown in Fig. \ref{GraphConstruction}, the main idea here is to generate dynamic filters from the predefined graphs $\mathcal{G}$ and the traffic embeddings $\mathcal{H}_i\in \mathcal{R}^{N \times d \times \tau_{i}}$ ($\tau_{i}$ is the sequence length at the $i$-th ST block), which are applied on the randomly initialized static node embeddings to construct dynamic adjacency matrices. In more detail, the core steps in Fig. \ref{GraphConstruction} are illustrated as follows:

\begin{figure}[htbp]
\centering
\includegraphics[width=350pt]{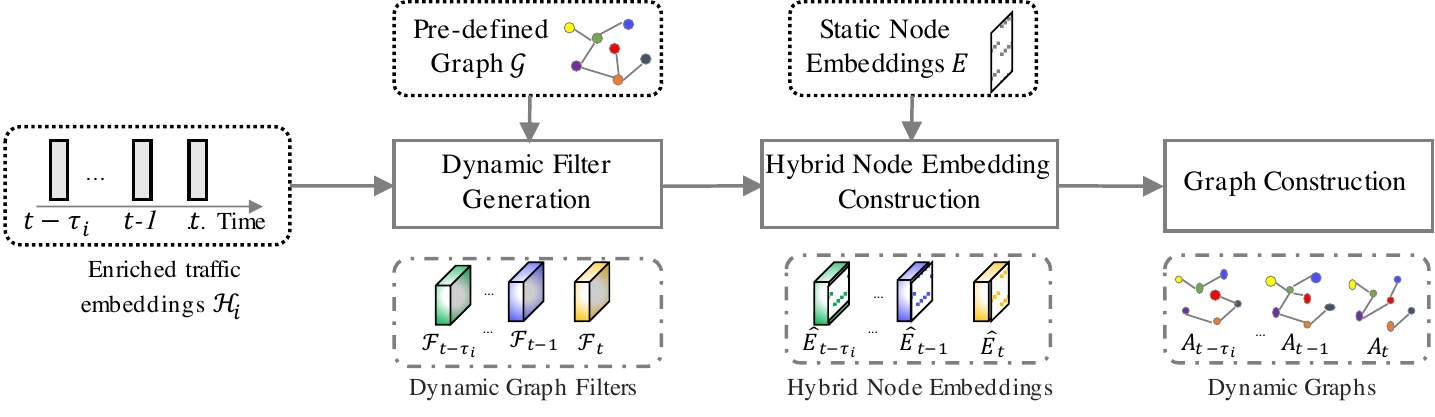}
\caption{Dynamic Graph Construction from the enriched traffic embeddings}
\label{GraphConstruction}
\vspace{-1em}
\end{figure}

\noindent \textbf{[Dynamic Filter Generation]} Given $\mathcal{H}_{i}=\{h_{t}\}\in \mathcal{R}^{N \times d \times \tau_{i}}$, the traffic embedding $h_{t}$ at time $t$ is firstly combined with the predefined adjacency matrix $A_{\mathcal{G}}\in \mathcal{R}^{N\times N}$ to generate dynamic graph filters via a diffusion convolution layer as proposed in \cite{li2018diffusion}: 
\begin{equation}
\small
    \mathcal{F}_{t}= \textstyle\sum_{k=0}^{K} P_{k} h_{t} W_{k}\in \mathcal{R}^{N\times d}
\end{equation}
where $K$ denotes the diffusion step, $P_{k}$= $A_{\mathcal{G}}$/$rowsum(A_{\mathcal{G}})$ represents the power series of the transition matrix \citep{wu2019graph}, and $W_{k}\in \mathcal{R}^{d\times d}$ is the model parameter matrix.

\noindent \textbf{[Hybrid Node Embedding Construction]}
Considering both the source and target traffic node, we initialize two random node embeddings $E^1,E^2 \in \mathcal{R}^{N\times d}$, representing the static node features \citep{wang2020traffic} which are not reflected in the observations but learnable during training. Thus, two dynamic filters are applied over the static node embeddings:
\begin{equation}
\small
    \begin{gathered}
    \hat{E}_{t}^1 = tanh(\alpha(\mathcal{F}_{t}^{1} \odot E^1)) \in \mathcal{R}^{N\times d}\\
    \hat{E}_{t}^2 = tanh(\alpha(\mathcal{F}_{t}^{1} \odot E^2)) \in \mathcal{R}^{N\times d}
  \end{gathered}
\end{equation}
where $\odot$ denotes the Hadamard product \citep{wu2019graph}. $\hat{E}_{t}^1$ and $\hat{E}_{t}^1$ are hybrid node embeddings combining both static and dynamic settings of the traffic data.

\noindent \textbf{[Graph Construction]}
As mentioned in the previous study \citep{wu2020connecting}, \textcolor{black}{\textit{in multivariate time series forecasting, we expect that the change of a node’s condition causes the change of another node’s condition such as traffic flow. Therefore the learned relationship is supposed to be uni-directional}}. We construct the graph by extracting uni-directional relationships between traffic nodes. The dynamic adjacency matrix is constructed from the hybrid embeddings:
\begin{equation}
\small
    A_{t} = ReLU(tanh(\alpha({\hat{E}_{t}^1\hat{E}_{t}^2}^T - {\hat{E}_{t}^2\hat{E}_{t}^1}^T))) \in \mathcal{R}^{N\times N}
\end{equation}
Therefore, we can construct the dynamic graphs $A_{\mathcal{D}_{i}}=\{A_{t}\}\in \mathcal{R}^{N \times N \times \tau_{i}}$for the enriched traffic embeddings $\mathcal{H}_i\in \mathcal{R}^{N \times d \times \tau_{i}}$ at the $i$-th ST block. \textcolor{black}{As the computation and memory cost grows quadratically with the increase of graph size, in practice, it is possible to adopt a sampling approach \citep{wu2020connecting}, which only calculates pairwise relationships among a subset of nodes.}

\subsection{Temporal Convolution Module}

\begin{figure}[htbp]
\centering
\includegraphics[width=350pt]{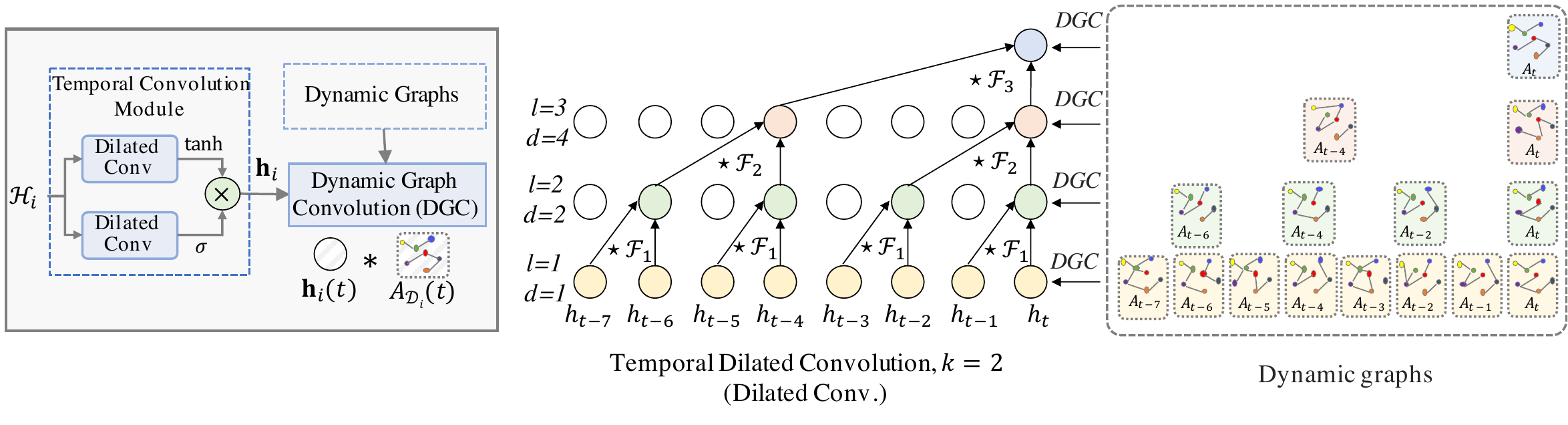}
\caption{Temporal Convolution module with Dynamic Graph Convolution}
\label{TCN}
\vspace{-1em}
\end{figure}

The temporal convolution network (TCN) \citep{lea2017temporal} consists of multiple dilated convolution layers, which allows extracting high-level temporal trends. Compared to RNN-based approaches, dilated causal convolution networks are capable of handling long-range sequences in a parallel manner. The output of the last layer is a representation that captures temporal dynamics in history. 
As shown in Fig. \ref{TCN}, considering the temporal dynamics in traffic data, we adopt the temporal convolution module \citep{wu2019graph} with the consideration of the gating mechanism over the enriched traffic embeddings $\mathcal{H}_i$. One dilated convolution block is followed by a tangent hyperbolic activation function to output the temporal features. The other block is followed by a sigmoid activation function as a gate to determine the ratio of information that can pass to the next module. \textcolor{black}{In particular, the sigmoid gate controls which input of the current states is relevant for discovering compositional structure and dynamic variances in time series. Applying the sigmoid nonlinearity on the input states differs from other well-known architectures (e.g., LSTM or GRU), which ignore the compositional structure features in time series \citep{yu2018spatio}}.

Given the enriched traffic embeddings $\mathcal{H}_i=\{h_{t}\}\in \mathcal{R}^{N \times d \times \tau_{i} }$, a filter $\mathcal{F}\in \mathcal{R}^{1 \times \mathrm{K}}$, $\mathrm{K}$ is the temporal filter size, $\mathrm{K}=2$ by default. The dilated causal convolution operation of $\mathcal{H}_i$ with $\mathcal{F}$ at time $t$ is represented as:
\begin{equation}
\small
    \mathcal{H}_{i} \star \mathcal{F}_{i}(t) = \textstyle\sum_{s=0}^{\mathrm{K}}\mathcal{F}_{i}(s) \mathcal{H}_{i}(t-\textbf{d} \times s) \in \mathcal{R}^{N \times d\times \tau_{i+1}}
\end{equation}
where $\star$ is the convolution operator, $\textbf{d}$ is the dilation factor, $d$ is the embedding dimension size, $\tau_{i+1}$ is the new sequence length after the convolution operation, which equals to one on the last layer. Fig. \ref{TCN} shows a three-layer dilated convolution block with $\mathrm{K}=2$, $\textbf{d}\in [1,2,4]$. 
Considering the gating mechanism, we define the output of the temporal convolution module:
\begin{equation}
\small
    \textbf{h}_i = tanh(W_{\mathcal{F}^{1}} \star \mathcal{H}_i) \odot \sigma(W_{\mathcal{F}^{2}} \star \mathcal{H}_i) \in \mathcal{R}^{N \times d \times \tau_{i+1}}
\end{equation}
where $W_{\mathcal{F}^{1}}$, $W_{\mathcal{F}^{2}}$ are learnable parameters of convolution filters, $\odot$ denotes the element-wise multiplication operator, $\sigma(\cdot)$ is the sigmoid function.

A classic temporal convolution module stacks the temporal features at each time step $t$. Therefore, the upper layer contains richer information than the lower layer. The gating mechanism allows filtering the temporal features on the lower layers by weighting features on different time steps without considering the spatial node interactions at each time step. Moreover, the spatial interactions in traffic data always show a dynamic nature \citep{wu2020connecting}. To this end, the gating mechanism from a dynamic spatial aspect is envisaged to better capture the Spatio-temporal patterns.

\subsection{Dynamic Graph Convolution}
Spatial interactions between the traffic nodes could be used to improve traffic forecasting performance. The dynamic spatial interaction leads to considering a dynamic version of graph convolution to conduct it on different graphs at different timestamps.
%Consider missing values
Different from previous work \citep{li2021dynamic} which uses raw traffic observations to mark the dynamic traffic status, we adopt the enriched traffic embeddings, which consider the missing-value issues to generate robust dynamic graphs.

As shown in Fig. \ref{TCN}, we apply the dynamic graph convolution on $\textbf{h}_i$, i.e., the output of the temporal convolution module, to further select the features at each time step from the spatial perspective.
As mentioned in Section \ref{Dynamic Graph Construction}, the dynamic graphs $A_{\mathcal{D}_{i}}\in \mathcal{R}^{N \times N \times \tau_{i}}$ are generated from the enriched traffic embeddings $\mathcal{H}_i\in \mathcal{R}^{N \times d \times \tau_{i}}$ at the $i$-th ST block. $A_t$ reflects the spatial relationships between nodes at time $t$. The temporal features $\textbf{h}_i(t)$ aggregate spatial information according to the adjacency matrix $A_t$. Inspired by DCRNN \citep{li2018diffusion}, we consider the traffic situation as the diffusion procedure on the graph. The graph convolution will generate the aggregated spatial information at each time step:
\begin{equation}
\small
    \mathcal{H}'_{i}(t) = \textstyle\sum_{k=0}^{K} \left ( A_{\mathcal{D}_{i}}(t) \right )^{k} \textbf{h}_{i}(t) W_{k}\in \mathcal{R}^{N\times d}
\end{equation}

\noindent where $K$ denotes the diffusion step, and $W_k$ is the learnable parameter matrix.
We adopt the residual connection \citep{he2016deep} between the input and output of each ST block to avoid the gradient vanishing issue in the model's training. Therefore, the input of the $(i+1)^{th}$ ST block is defined as:
\begin{equation}
\small
    \mathcal{H}_{i+1}(t) = \mathcal{H}_{i}(t) + \mathcal{H}'_{i}(t) 
\end{equation}

\subsection{Output Forecasting Module}
% the skip connection layer
% optimisation, i.e. loss function
The outputs $\textbf{h}_{i} \in \mathcal{R}^{N \times d \times \tau_{i+1} }$ of the middle temporal convolution modules and $\mathcal{H}_{l}\in \mathcal{R}^{N \times d \times 1 }$ of the last ST block are considered for the final prediction, which represent the hidden states at various Spatio-temporal levels. We add skip connections on each of the hidden states which are essentially $1 \times \tau_{i+1}$ standard convolutions ( $\tau_{i+1}$ denotes the sequence length at the output of the $i$-th ST block). The concatenated output features are defined as follows:
\begin{equation}
\small
    O = (\textbf{h}_{0} W_{s}^{0} + b_{s}^{0})\| ... \| (\textbf{h}_{i} W_{s}^{i} + b_{s}^{i}) \| ... \| (\textbf{h}_{l-1} W_{s}^{l-1} + b_{s}^{l-1}) \ \| (\mathcal{H}_{l} W_{s}^{l} + b_{s}^{l})
\end{equation}
where $O \in \mathcal{R}^{N\times (l+1)d}$, $W_{s}^{i}$, $b_{s}^{i}$ are learnable parameters for the convolutions. Two fully-connected layers are added to project the concatenated features into the desired output dimension:
\begin{equation}
\small
    \hat{\textbf{Y}} = (ReLU(OW_{fc}^{1} +  b_{fc}^{1})) W_{fc}^{2}  +  b_{fc}^{2} \in \mathcal{R}^{N\times T_{p}}
\end{equation}
where $W_{fc}^{1}$, $W_{fc}^{2}$, $b_{fc}^{1}$, $b_{fc}^{2}$ are learnable parameters for the fully-connected layers, $N$ is the node number, $T_p$ denotes the forecasting steps.

Given the ground truth $\mathbf{Y}\in \mathcal{R}^{N\times T_{p}}$ and the predictions $\hat{\textbf{Y}}\in \mathcal{R}^{N\times T_{p}}$, we use mean absolute error (MAE) as our model's loss function for training:
\begin{equation}
\small
L = \frac{1}{NT_{p}}\textstyle\sum_{n=1}^{N}\textstyle\sum_{t=1}^{T_{p}} \lvert \hat{\textbf{Y}}_{t}^{n} - \textbf{Y}_{t}^{n}  \rvert
\end{equation}
%\textcolor{black}{We should note that the missing values are masked out when computing the loss error.} 

\section{Experiments}
In this section, we demonstrate the effectiveness of GCN-M \footnote{The source code is publicly available in https://github.com/JingweiZuo/GCN-M} with real-life traffic datasets. The experiments were designed to answer the following research questions (RQs):
\begin{itemize}[leftmargin=.5in]

    \item[RQ 1] \textit{Performance on raw benchmark datasets:} \textcolor{black}{How well does our model perform on traffic datasets with a few missing values or without?} 
    \item[RQ 2] \textit{Complex scenarios of missing values:} How successful is our model at forecasting traffic data considering the complex missing values scenarios? % parameter effects, i.e., L, S; Ablation study 
    \item[RQ 3] \textit{Dynamic graph modeling:} How does our method perform on dynamic graph modeling considering the missing values?  %different strategy for generating the dynamic graphs: raw observation, enriched traffic embeddings, remove dynamic graph (i.e., static graph, predefined graph)
    \textcolor{black}{\item[RQ 4] \textit{One-step processing VS two-step processing:} How will our method perform when adopting distinct missing-value processing strategies? }
    %\item[RQ 5] \textit{Missing values OR zero values:} How well can our model identify the missing measures from the inherent zero values?
    
\end{itemize}

\subsection{Experimental settings}
\noindent \textbf{[Datasets]} 
We base our experiments on the public traffic datasets: PEMS-BAY and METR-LA released by \cite{li2018diffusion}, which are widely used in the literature. PEMS-BAY records six months of traffic speed on 325 sensors in the California Bay Area. METR-LA records four months of traffic flow on 207 sensors on the highways of Los Angeles County. Both datasets contain some zero and/or missing values, though PEMS-BAY has been pre-processed by the domain experts from the data provider \citep{PEMS_manual} to interpolate most of the missing values. Following \cite{li2018diffusion}, the datasets are split with 70\% for training, 10\% for validation, and 20\% for testing. \textcolor{black}{In order to validate the model in complex scenarios of missing values, we introduce complex missing values in the datasets (see details in Section \ref{section_rq2}). In practice, the model should forecast future values from the input data with missing values. Therefore, in the testing set, we mask out the observations from the input sequence $\mathcal{X}$ (i.e., inject missing values) but maintain the complete information for the target $\mathbf{Y}$.} We use recent $\tau$= 12 timestamps as input to predict the next $T_p$ timestamp. Considering that the missing values are marked as zeros, we scale the input by dividing it with the max speed of the training set instead of applying Z-score normalization. \textcolor{black}{This avoids changing the zero values and facilitates the computation process.}
Table \ref{TrafficDatasets} shows the summary statistics of the datasets.
\begin{table}[!htbp]
\centering
\caption{Summary statistics of PEMS-BAY and METR-LA}
\label{TrafficDatasets}
\scalebox{0.85}{
\begin{tabular}{ccccccc}
\toprule
\textbf{Data} & \textbf{\#Nodes} & \textbf{\#Edges} & \textbf{Length} & \textbf{Sample Rate} & \textbf{Observations} & \textbf{Zero ratio}  \\
\midrule
PEMS-BAY      & 325              & 2369             & 52 116          & 5 mins               & 16 937 179            & 0.0031\%             \\
METR-LA       & 207              & 1515             & 34 272          & 5 mins               & 6 519 002             & 8.11\% \\
\bottomrule
\end{tabular}
}
\end{table}

\noindent \textbf{[Evaluation metrics]}
The forecasting accuracy of all tested models is evaluated by three metrics: mean absolute error (MAE), root mean square error (RMSE) and mean absolute percentage error (MAPE).
\begin{equation}
\small
\begin{gathered}
    MAE(\textbf{Y}, \hat{\textbf{Y}}) = \frac{1}{NT_{p}}\textstyle\sum_{n=1}^{N}\textstyle\sum_{t=1}^{T_{p}} \lvert \hat{\textbf{Y}}_{t}^{n} - \textbf{Y}_{t}^{n}  \rvert\\
    RMSE(\textbf{Y}, \hat{\textbf{Y}}) = \sqrt{\frac{1}{NT_{p}}\textstyle\sum_{n=1}^{N}\textstyle\sum_{t=1}^{T_{p}} \lvert \hat{\textbf{Y}}_{t}^{n} - \textbf{Y}_{t}^{n}  \rvert^{2} }\\
    MAPE(\textbf{Y}, \hat{\textbf{Y}}) = \frac{1}{NT_{p}}\textstyle\sum_{n=1}^{N}\textstyle\sum_{t=1}^{T_{p}} \lvert \frac{\hat{\textbf{Y}}_{t}^{n} - \textbf{Y}_{t}^{n}}{ \textbf{Y}_{t}^{n}}  \rvert
  \end{gathered}
\end{equation}
and $N$ denotes the node numbers, $T_p$ represents the forecasting steps. 

\textcolor{black}{When evaluating each model's performance on the testing set, we mask out the inherent zero values in the prediction targets when computing the metrics. } %When evaluating the models in different horizons, we only report the metrics on single step. 

\textcolor{black}{We conduct statistical tests to assess the statistical significance of the differences between the models. In order to compare the forecasters over multiple datasets \citep{IsmailFawaz2018deep}, we adopted the critical difference diagrams recommended by \cite{demsar2006statistical} and used the Friedman test \citep{friedman1940comparison} to reject the null hypothesis (i.e., check whether there are significant differences at all). We followed the pairwise post-hoc analysis recommended by \cite{benavoli2016should} and adapted the critical difference diagrams \citep{demsar2006statistical} with the change that all forecasters are compared with pairwise Wilcoxon signed-rank test \citep{wilcoxon1992individual}. Additionally, we formed cliques using Holm's alpha (5\%) correction \citep{holm1979simple} rather than the post-hoc Nemenyi test originally used in \cite{demsar2006statistical}.
}

\noindent \textbf{[Execution and Parameter Settings]}
The proposed model is implemented by PyTorch 1.6.0 and is trained using the Adam optimizer with a learning rate of 0.001. All the models are tested on a single Tesla V100 GPU of 32 Go memory. In the multi-scale memory module, $L$, $S$ are set to 12 and 5. $n_h$, $n_d$, $n_w$ are all set to 2. We apply four ST blocks in which the Temporal Convolution module contains two dilated layers with dilation factor $\textbf{d} \in \left[1,2\right]$. The embedding dimension $d$ is set to 32.

\subsection{Baseline Approaches}
We only compare with the baseline models whose source code is publicly available. We follow the default parameter settings described in each
paper for training each model. According to the strategy for handling missing values, the baseline models can be organized into two categories: 
%including both missing-value processing approaches and specific forecasting models considering missing values. We adopt the default parameter settings described in each paper for testing. The methods are summarized as follows: 
\begin{enumerate}
    \item \textcolor{black}{Ignore the missing values when optimizing the model, i.e., consider missing values in the input sequence as actual zero values and mask out the missing values when computing the loss error:}
    \begin{itemize}
        \item DCRNN \citep{li2018diffusion}: Based on the predefined graphs, DCRNN integrates GRU with dual directional diffusion convolution.
        \item STGCN \citep{yu2018spatio}: Based on the predefined graphs, STGCN combines graph convolution into 1D convolutions.
        \item Graph WaveNet \citep{wu2019graph}: Graph WaveNet learns an adaptive graph and integrates diffusion graph convolutions with temporal convolutions.
        \item MTGNN \citep{wu2020connecting}: MTGNN learns an adaptive graph and integrates mix-hop propagation layers in the graph convolution module. Moreover, it designed the dilated inception layers in temporal convolutions.
        \item AGCRN \citep{bai2020adaptive}: AGCRN learns an adaptive graph and integrates with recurrent graph convolutions with node adaptive parameter learning.
        \item GTS \citep{shang2020discrete}: GTS learns a probabilistic graph which is combined with the recurrent graph convolutions to do traffic forecasting.
        \begin{description}
         \item[Note:] In practice, any model can ignore the missing values in their optimization process. We list here some classic models and the most recent models designed specifically for traffic forecasting.
         \end{description}
    \end{itemize}

   \item Jointly model the missing values and forecasting task, i.e., one-step processing models:
   \begin{itemize}
     \item GRU \citep{chung2014empirical}: Gated Recurrent Unit (GRU) can be considered as a basic structure for traffic forecasting.
     \item GRU-I \citep{che2018recurrent}: A variation of GRU, which infers the missing values with the predictions from previous steps.
     \item GRU-D \citep{che2018recurrent}: Based on GRU, GRU-D helps improve the prediction performance by incorporating the missing patterns, including the masking information and time intervals between missing and observed values.
     \item LSTM-I \citep{cui2020graph}: Based on LSTM, LSTM-I is similar to GRU-I for inferring the missing values.
     \item LSTM-M \citep{tian2018lstm}: Based on LSTM, LSTM-M is designed for traffic forecasting on data with short-period and long-period missing values.
     \item SGMN \citep{cui2020graph}: Based on the graph Markov process, SGMN does traffic forecasting on data with random missing values by corporating a spectral graph convolution.
     %\begin{description}
     %\item[Note:] I would like to describe something here
     %\end{description}
   \end{itemize}
   
    %\item Preprocess the missing values before training the traffic forecasting models, i.e., two-step processing
    %\begin{itemize}
    %    \item Any of the models as mentioned above can apply to this category.
    %\end{itemize}
\end{enumerate}

\subsection{RQ 1: Performance on raw benchmark datasets}
Recently, a lot of traffic forecasting models \citep{jiang2022graph} have been proposed, achieving remarkable performance on the benchmark datasets PEMS-BAY and METR-LA. Our objective is not to beat all the models in terms of forecasting accuracy, but to validate our proposal for jointly modeling missing values and forecasting. Therefore, it's essential to know how GCN-M performs in a primary setting, i.e., on the original datasets with a few missing values or without.
We pick three classic models (DCRNN \citep{li2018diffusion}, STGCN \citep{yu2018spatio} and Graph WaveNet \citep{wu2019graph}) and the three most recent models (MTGNN \citep{wu2020connecting}, AGCRN \citep{bai2020adaptive} and GTS \citep{shang2020discrete}), which focus on the Spatio-temporal modeling of traffic data, and generally ignore the missing values when training the model. Additionally, we consider the group of works \citep{che2018recurrent,cui2020graph,tian2018lstm} which are specifically designed for modeling the missing values in the forecasting model, i.e., one-step processing models. 

\begin{table}[!htbp]
\centering
\caption{Performance comparison on \textbf{raw} PEMS-BAY dataset. \textcolor{black}{The results show the forecasting errors at each forecasting step (i.e., horizon).}}
\label{BAY_complete}
\scalebox{0.68}{
\begin{tabular}{ccccccccccccc}
\toprule
\textbf{PEMS-BAY} & \multicolumn{3}{c}{Horizon=1 (5 mins)}          & \multicolumn{3}{c}{Horizon=3 (15 mins)}         & \multicolumn{3}{c}{Horizon=6 (30 mins)}         & \multicolumn{3}{c}{Horizon=12 (60 mins)}         \\

Models            & MAE           & RMSE          & MAPE            & MAE           & RMSE          & MAPE            & MAE           & RMSE          & MAPE            & MAE           & RMSE          & MAPE             \\
\midrule
DCRNN             & 0.96          & 1.63          & 1.81\%          & 1.38          & 2.95          & 2.90\%          & 1.74          & 3.97          & 3.90\%          & 2.07          & 4.74          & 4.90\%           \\
STGCN             & 0.98          & 1.84          & 1.98\%          & 1.44          & 2.88          & 3.16\%          & 1.85          & 3.82          & 4.20\%          & 2.21          & 4.52          & 5.09\%           \\
GraphWaveNet      & 0.91          & \textbf{1.56} & 1.72\%          & \textbf{1.31} & 2.75          & \textbf{2.73\%} & 1.65          & 3.75          & 3.74\%          & 1.99          & 4.62          & 4.78\%           \\
MTGNN             & \textbf{0.87} & 1.57          & \textbf{1.70\%} & 1.33          & 2.80          & 2.80\%          & 1.65          & 3.75          & 3.70\%          & 1.95          & 4.49          & 4.56\%           \\
AGCRN             & 0.95          & 1.81          & 1.94\%          & 1.37          & 2.92          & 2.94\%          & 1.69          & 3.87          & 3.82\%          & 1.99          & 4.61          & 4.62\%           \\
GTS               & 0.91          & 1.64          & 1.77\%          & 1.32          & 2.80          & 2.75\%          & 1.63          & 3.74          & \textbf{3.63\%} & \textbf{1.90} & \textbf{4.40} & \textbf{4.44\%}  \\ 
\cmidrule{1-1}
GRU               & 1.29          & 2.46          & 2.54\%          & 1.89          & 3.53          & 3.98\%          & 2.27          & 4.24          & 5.02\%          & 2.65          & 4.90          & 5.92\%           \\
GRU-I             & 1.30          & 2.57          & 2.57\%          & 1.89          & 3.52          & 3.99\%          & 2.26          & 4.22          & 4.99\%          & 2.62          & 4.89          & 5.87\%           \\
GRU-D             & 5.40          & 9.25          & 13.83\%         & 5.34          & 9.25          & 13.76\%         & 5.42          & 9.26          & 13.85\%         & 5.41          & 9.27          & 13.85\%          \\
LSTM-I            & 1.71          & 2.69          & 2.80\%          & 1.97          & 3.45          & 4.08\%          & 2.57          & 5.52          & 5.62\%          & 2.74          & 5.00          & 6.21\%           \\
LSTM-M            & 1.35          & 2.31          & 2.71\%          & 1.87          & 3.39          & 3.95\%          & 2.33          & 4.33          & 5.17\%          & 3.45          & 8.32          & 7.29\%           \\
SGMN              & 0.98          & 1.85          & 1.88\%          & 1.63          & 3.40          & 3.32\%          & 2.29          & 4.91          & 4.88\%          & 3.31          & 6.86          & 7.32\%           \\
\textbf{GCN-M (ours)}      & 0.91          & 1.57          & 1.75\%          & 1.33          & \textbf{2.72} & 2.76\%          & \textbf{1.62} & \textbf{3.64} & 3.64\%          & 1.95          & \textbf{4.40} & 4.61\%      \\ 
\bottomrule
\end{tabular}
}
\end{table}

\begin{table}[!htbp]
\centering
\caption{Performance comparison on \textbf{raw} METR-LA dataset. \textcolor{black}{The results show the forecasting errors at each forecasting step (i.e., horizon).}}
\label{LA_complete}
\scalebox{0.68}{
\begin{tabular}{ccccccccccccc}
\toprule
\textbf{METR-LA} & \multicolumn{3}{c}{Horizon=1 (5 mins)}          & \multicolumn{3}{c}{Horizon=3 (15 mins)}         & \multicolumn{3}{c}{Horizon=6 (30 mins)}         & \multicolumn{3}{c}{Horizon=12 (60 mins)}         \\

Models            & MAE           & RMSE          & MAPE            & MAE           & RMSE          & MAPE            & MAE           & RMSE          & MAPE            & MAE           & RMSE          & MAPE             \\
\midrule
DCRNN                 & 2.45          & 4.21          & 5.99\%          & 2.77          & 5.38          & 7.30\%          & 3.15          & 6.45  & 8.80\%          & 3.60          & 7.60          & 10.50\%          \\
STGCN                 & 2.58          & 4.32          & 6.22\%          & 3.04          & 5.48          & 8.00\%          & 3.60          & 6.51  & 9.97\%          & 4.21          & 7.37          & 11.61\%          \\
GraphWaveNet          & 2.41          & 4.29          & 5.93\%          & \textbf{2.68} & \textbf{5.14} & 6.87\%          & 3.06          & 6.14  & 8.23\%          & 3.52          & 7.25          & \textbf{9.77\%}  \\
MTGNN                 & \textbf{2.24} & 3.92          & \textbf{5.39\%} & \textbf{2.68} & 5.16          & \textbf{6.86\%} & \textbf{3.05} & 6.16  & 8.19\%          & \textbf{3.50} & \textbf{7.24} & 9.83\%           \\
AGCRN                 & 2.41          & 4.27          & 6.08\%          & 2.86          & 5.54          & 7.66\%          & 3.22          & 6.55  & 8.92\%          & 3.58          & 7.45          & 10.24\%          \\
GTS                   & 2.32          & 4.15          & 6.12\%          & 2.72          & 5.42          & 7.11\%          & 3.11          & 6.47  & \textbf{7.49\%} & 3.52          & 7.49          & 10.07\%          \\
\cmidrule{1-1}
GRU                   & 2.83          & 4.56          & 6.78\%          & 3.48          & 5.80          & 9.02\%          & 3.97          & 6.74  & 10.72\%         & 4.65          & 7.86          & 13.00\%          \\
GRU-I                 & 2.80          & 4.52          & 6.70\%          & 3.49          & 5.83          & 9.05\%          & 3.97          & 6.74  & 10.75\%         & 4.60          & 7.88          & 12.80\%          \\
GRU-D                 & 7.46          & 11.82         & 24.55\%         & 7.43          & 11.85         & 24.62\%         & 7.45          & 11.84 & 24.62\%         & 7.47          & 11.86         & 24.68\%          \\
LSTM-I                & 2.86          & 4.57          & 6.77\%          & 3.57          & 5.88          & 9.05\%          & 4.10          & 6.85  & 10.94\%         & 4.78          & 8.13          & 13.34\%          \\
LSTM-M                & 3.15          & 5.58          & 7.03\%          & 3.46          & 5.74          & 8.75\%          & 4.08          & 6.86  & 10.89\%         & 4.63          & 7.83          & 12.92\%          \\
SGMN                  & 3.11          & 6.02          & 7.01\%          & 4.23          & 8.54          & 9.89\%          & 5.46          & 10.88 & 13.01\%         & 7.37          & 13.78         & 17.81\%          \\
\textbf{GCN-M (ours)} & 2.34          & \textbf{3.89} & 5.88\%          & 2.74          & 5.21          & 6.94\%          & 3.12          & 6.18  & 8.25\%          & 3.54          & \textbf{7.12} & 10.01\%     \\ 
\bottomrule
\end{tabular}
}
\end{table}

\begin{figure*}[htbp]

\centering
  \subfloat[][\centering MAE]{
  \hspace{-1cm}
        \includegraphics[width=0.55\linewidth]{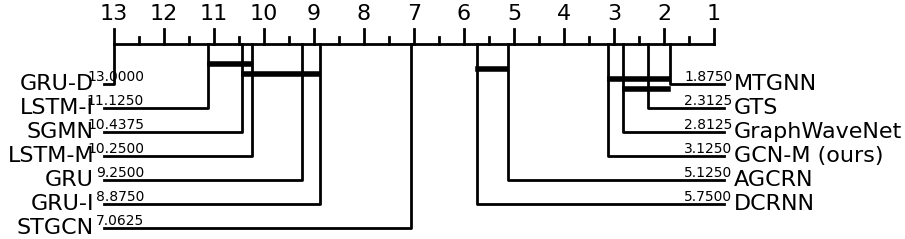}\label{fig_raw_CD_diag:mae}}
  \subfloat[][\centering RMSE
  \hspace{-1cm}
  \label{fig_raw_CD_diag:rmse}]{%
        \includegraphics[width=0.55\linewidth]{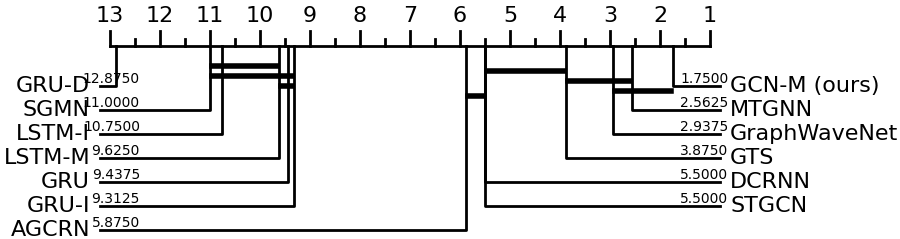}}
        \\
  \subfloat[][\centering MAPE \label{fig_raw_CD_diag:mape}]{%
        \includegraphics[width=0.55\linewidth]{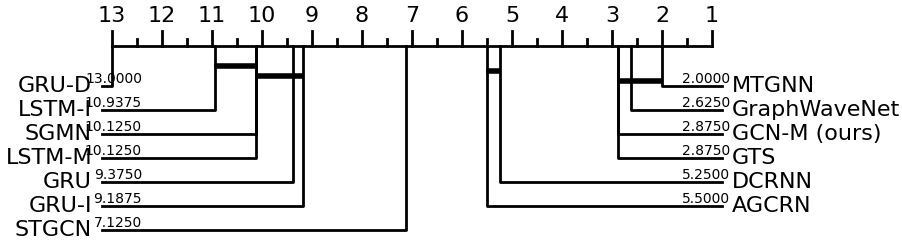}}
    \hfill
  
  \caption{Critical difference diagrams showing rankings by various evaluation metrics and the statistical difference comparison of 13 forecasting models on PEMS-BAY and METR-LA datasets under multiple forecasting horizons.}
  \label{fig_raw_CD_diag}
  \vspace{-1.5em}
\end{figure*}

Table \ref{BAY_complete} and \ref{LA_complete} show the performance comparison on the raw PEMS-BAY and METR-LA datasets, respectively. It should be noted that the original datasets already contain missing values (0.0031\% missed in PEMS-BAY, 8.11\% missed in METR-LA). We train the models for single-step (horizon=1) and multi-step (horizon =3,6,12) forecasting. We report the evaluation errors on each horizon step. We observe from the results that no model achieved evident better performance than the others. However, the first group of works (e.g., DCRNN) performs better than the one-step processing models, which is not surprising as they incorporate the advanced graph models (e.g., \textit{mix-hop propagation} \citep{wu2020connecting}) and training techniques (e.g., \textit{curriculum learning} \citep{wu2020connecting}) to improve the Spatio-temporal forecasting performance. Surprisingly, among the one-step processing models, GRU-D \citep{che2018recurrent} shows much worse performance than the others, probably due to the fact that it has been designed for health care applications, whose data is more stable than dynamic traffic data. LSTM-M \citep{tian2018lstm} and SGMN \citep{cui2020graph}, designed for traffic forecasting with missing values, show relatively good performance in PEMS-BAY especially on single-step forecasting. However, they did not show a clear advantage over the first group of works. The one-step processing models are generally designed for single-step forecasting; their performance gap with the first group of works becomes larger under a multi-step forecasting setting. 

\textcolor{black}{We present in Figure \ref{fig_raw_CD_diag} the 
critical difference diagrams \citep{demsar2006statistical} which show the average rankings and visualize the statistical difference between the forecasting models, where a thick horizontal line shows a group (i.e., clique) of models that are not significantly different in terms of evaluation metrics. From Figure \ref{fig_raw_CD_diag}, we observe that even though GCN-M belongs to the one-step processing models, its performance remains close to the first group of works. GCN-M is not significantly different to MTGNN, GTS, and GraphWaveNet on all three evaluation metrics. 
}
Moreover, the advanced graph models and training techniques in recent work \citep{wu2020connecting} can be considered to have improved the performance of GCN-M further.

\subsection{RQ 2: Complex scenarios of missing values}\label{section_rq2}
In this section, we demonstrate the power of GCN-M in handling complex scenarios of missing values for the purpose of traffic forecasting. 

As mentioned previously in Figure \ref{MissingValueScenarios}, there are several scenarios of missing values in real-life traffic datasets (e.g., METR-LA): short-range or long-range missing; partial or entire network missing. The results in Table \ref{BAY_complete} and \ref{LA_complete} did not show the superiority of GCN-M over other models on the original datasets with a low missing rate. To test the model's capability of handling complex missing values, we have designed three scenarios with various missing rates (10\%, 20\%, and 40\%), and removed the observations from the datasets accordingly. \textcolor{black}{We use $\hat{x}_{i} \in \mathcal{R}^{n \times f \times t}$ to represent each of the observation tensors to be removed from $\mathcal{X} \in \mathcal{R}^{N \times F \times \tau}$, therefore a local mask tensor $\hat{m}_{i} \in \mathcal{R}^{n \times f \times t} $ can be defined accordingly. This annotates the locations of missing values in the original dataset. All the local mask tensors constitute the global mask sequence $\mathcal{M} = \{\hat{m}_{i}\}$ which allows injecting missing values with a given missing rate.} Then, we designed the scenarios of:
\begin{itemize}
    \item Short-range missing: we randomly set $n\in [1,...,N]$, $f=F$, $t=1$
    \item Long-range missing: we randomly set $n\in [1,...,N]$, $f=F$, $t=\tau$
    \item Mix-range missing: we randomly set $n\in [1,...,N]$, $f=F$, $t \in [1,...,\tau]$
\end{itemize}

\begin{table}[t]
\centering
\caption{Performance comparison on the \textbf{incomplete} PEMS-BAY dataset with various random settings on missing values. The results show a one-hour (12-step) average of the forecasting errors. The underlined values represent the best results among the one-step processing models, the bold values represent the best results among all the models.}
\label{BAY_combine_miss}
\scalebox{0.72}{
\begin{tabular}{cllllllllll}
\toprule
                &       & \multicolumn{3}{c}{Missing Rate = 10\%}                                   & \multicolumn{3}{c}{Missing Rate = 20\%}                                    & \multicolumn{3}{c}{Missing Rate = 40\%}                                  \\
          & Models                & MAE                 & RMSE          & MAPE        & MAE                 & RMSE          & MAPE       & MAE                 & RMSE          & MAPE         \\
\midrule
\multirow{12}{*}{\rotatebox{90}{Short-range missing}} & DCRNN                 & 1.76                  & 3.94                  & 3.94\%                  & 1.82                  & 3.96                  & 4.01\%                  & 1.85                  & 4.26                  & 4.04\%                   \\
                    & STGCN                 & 1.82                  & 4.11                  & 4.25\%                  & 1.91                  & 4.18                  & 4.41\%                  & 1.97                  & 4.33                  & 4.42\%                   \\
                    & GraphWaveNet          & 1.69                  & 3.79                  & 3.81\%                  & 1.74                  & 3.75                  & 3.75\%                  & 1.79                  & 3.87                  & 3.90\%                   \\
                    & MTGNN                 & \textbf{1.58}         & \textbf{3.42}         & \textbf{3.33\%}         & 1.72                  & 3.78                  & 3.83\%                  & 1.83                  & 4.03                  & 3.94\%                   \\
                    & AGCRN                 & 1.65                  & 3.81                  & 3.78\%                  & 1.66                  & 3.81                  & 3.79\%                  & 1.72                  & 3.96                  & 3.95\%                   \\
                    & GTS                   & 1.65                  & 3.86                  & 3.74\%                  & 1.65                  & 3.86                  & 3.76\%                  & \textbf{1.69}         & 3.92                  & \textbf{3.86\%}          \\
\cmidrule{2-2}
                    & GRU                   & 2.60                  & 4.64                  & 5.75\%                  & 2.67                  & 4.78                  & 5.90\%                  & 2.86                  & 5.10                  & 6.37\%                   \\
                    & GRU-I                 & 2.29                  & 4.28                  & 5.06\%                  & 2.31                  & 4.31                  & 5.09\%                  & 2.41                  & 4.47                  & 5.38\%                   \\
                    & GRU-D                 & 5.38                  & 9.29                  & 13.84\%                 & 5.46                  & 9.36                  & 13.96\%                 & 7.20                  & 11.58                 & 16.91\%                  \\
                    & LSTM-I                & 2.35                  & 4.33                  & 5.22\%                  & 2.82                  & 6.63                  & 6.05\%                  & 3.06                  & 7.47                  & 6.56\%                   \\
                    & LSTM-M                & 2.47                  & 4.55                  & 5.50\%                  & 2.56                  & 4.70                  & 5.74\%                  & 3.34                  & 7.09                  & 7.68\%                   \\
                    & SGMN                  & 2.32                  & 4.96                  & 4.94\%                  & 2.34                  & 5.01                  & 5.00\%                  & 2.45                  & 5.20                  & 5.23\%                   \\
                    & \textbf{GCN-M (ours)} & \uline{1.62}          & \uline{3.67}          & \uline{3.60\%}          & \textbf{\uline{1.63}} & \textbf{\uline{3.73}} & \textbf{\uline{3.68\%}} & \uline{1.75}          & \textbf{\uline{3.81}} & \uline{3.90\%}           \\

\midrule
\multirow{12}{*}{\rotatebox{90}{Long-range missing}} & DCRNN                 & 1.83                & 4.07          & 4.22\%          & 1.96                & 4.22          & 4.42\%          & 2.07                & 4.45          & 4.67\%           \\
                    & STGCN                 & 1.92                & 4.22          & 4.42\%          & 2.03                & 4.37          & 4.72\%          & 2.14                & 4.52          & 4.76\%           \\
                    & GraphWaveNet          & 1.74                & 3.96          & 4.03\%          & 1.87                & 4.09          & 4.18\%          & 1.94                & 4.21          & 4.33\%           \\
                    & MTGNN                 & \textbf{1.65}       & \textbf{3.68} & \textbf{3.72\%} & 1.89                & 4.01          & 4.17\%          & 2.01                & 4.42          & 4.61\%           \\
                    & AGCRN                 & 1.72                & 3.78          & 3.94\%          & 1.84                & 4.11          & 4.13\%          & 1.90                & 4.18          & 4.31\%           \\
                    & GTS                   & 1.68                & 3.86          & 3.91\%          & 1.78                & 4.12          & 4.97\%          & 1.88                & 4.17          & 4.22\%           \\
\cmidrule{2-2}
                    & GRU                   & 2.93                & 5.12          & 6.32\%          & 3.06                & 5.31          & 6.63\%          & 3.35                & 5.78          & 7.03\%           \\
                    & GRU-I                 & 2.52                & 4.51          & 5.33\%          & 2.53                & 4.57          & 5.73\%          & 2.71                & 4.82          & 5.51\%           \\
                    & GRU-D                 & 9.33                & 14.51         & 22.31\%         & 9.89                & 13.94         & 22.86\%         & 11.07               & 15.88         & 23.13\%          \\
                    & LSTM-I                & 2.65                & 4.65          & 5.88\%          & 3.13                & 6.35          & 6.82\%          & 3.62                & 9.53          & 7.12\%           \\
                    & LSTM-M                & 3.93                & 7.25          & 9.17\%          & 5.45                & 10.06         & 13.67\%         & 5.57                & 10.12         & 14.59\%          \\
                    & SGMN                  & 8.86                & 12.57         & 14.54\%         & 11.45               & 14.56         & 18.31\%         & 14.62               & 17.23         & 23.13\%          \\
                    & \textbf{GCN-M (ours)} & \uline{1.70}          & \uline{3.75}          & \uline{3.74\%}          & \textbf{\uline{1.73}} & \textbf{\uline{3.88}} & \textbf{\uline{3.92\%}} & \textbf{\uline{1.79}} & \textbf{\uline{4.07}} & \textbf{\uline{4.14\%}}  \\

\midrule
\multirow{12}{*}{\rotatebox{90}{Mix-range missing}}   & DCRNN                 & 1.81                & 4.01          & 4.15\%          & 1.91                & 4.16          & 4.31\%          & 2.02                & 4.36          & 4.52\%           \\
                    & STGCN                 & 1.85                & 4.13          & 4.21\%          & 1.98                & 4.31          & 4.56\%          & 2.11                & 4.43          & 4.68\%           \\
                    & GraphWaveNet          & 1.72                & 3.92          & 3.96\%          & 1.83                & 4.06          & 4.14\%          & 1.89                & 4.11          & 4.21\%           \\
                    & MTGNN                 & 1.69                & 3.77          & 3.78\%          & 1.86                & 4.03          & 4.11\%          & 1.98                & 4.32          & 4.44\%           \\
                    & AGCRN                 & 1.67                & 3.85          & 3.88\%          & 1.72                & 3.95          & 3.99\%          & 1.80                & 4.10          & 4.13\%           \\
                    & GTS                   & 1.70                & 3.96          & 3.92\%          & 1.75                & 3.98          & 3.89\%          & 1.79                & 4.09          & 4.09\%           \\
\cmidrule{2-2}
                    & GRU                   & 2.71                & 4.88          & 6.03\%          & 2.82                & 5.08          & 6.28\%          & 3.05                & 5.43          & 6.82\%           \\
                    & GRU-I                 & 2.31                & 4.30          & 5.11\%          & 2.34                & 4.39          & 5.18\%          & 2.40                & 4.50          & 5.37\%           \\
                    & GRU-D                 & 8.90                & 13.71         & 20.03\%         & 9.46                & 14.50         & 21.04\%         & 10.21               & 15.19         & 22.44\%          \\
                    & LSTM-I                & 2.46                & 4.51          & 5.49\%          & 2.75                & 5.85          & 6.02\%          & 3.39                & 9.15          & 6.88\%           \\
                    & LSTM-M                & 3.86                & 7.06          & 8.93\%          & 5.19                & 9.71          & 13.15\%         & 5.27                & 9.74          & 13.29\%          \\
                    & SGMN                  & 7.41                & 10.91         & 13.47\%         & 9.95                & 13.49         & 17.56\%         & 13.10               & 16.96         & 22.58\%          \\
                    & \textbf{GCN-M (ours)} & \textbf{\uline{1.65}} & \textbf{\uline{3.67}} & \textbf{\uline{3.69\%}} & \textbf{\uline{1.66}} & \textbf{\uline{3.72}} & \textbf{\uline{3.62\%}} & \textbf{\uline{1.69}} & \textbf{\uline{3.79}} & \textbf{\uline{3.83\%}} \\
\bottomrule
\end{tabular}
}
\end{table}

\begin{table}[t]
\centering
\caption{Performance comparison on the \textbf{incomplete} METR-LA dataset with various random settings on missing values. The results show a one-hour (12-step) average of the forecasting errors. The underlined values represent the best results among the one-step processing models, the bold values represent the best results among all the models.}
\label{LA_combine_miss}
\scalebox{0.72}{
\begin{tabular}{cllllllllll}
\toprule
                &       & \multicolumn{3}{c}{Missing Rate = 10\%}                                   & \multicolumn{3}{c}{Missing Rate = 20\%}                                    & \multicolumn{3}{c}{Missing Rate = 40\%}                                  \\
          & Models                & MAE                 & RMSE          & MAPE        & MAE                 & RMSE          & MAPEy       & MAE                 & RMSE          & MAPE         \\
\midrule
\multirow{12}{*}{\rotatebox{90}{Short-range missing}} & DCRNN        & 3.31                  & 6.61                & 9.47\%          & 3.44                  & 6.80                  & 9.57\%                  & 3.50                  & 6.90                  & 9.68\%                   \\
                    & STGCN        & 3.53                  & 7.08                & 9.73\%          & 3.59                  & 7.25                  & 10.25\%                 & 3.66                  & 7.45                  & 10.41\%                  \\
                    & GraphWaveNet & 3.28                  & 6.60                & 9.11\%          & 3.36                  & 6.74                  & 9.50\%                  & 3.45                  & 6.81                  & 9.57\%                   \\
                    & MTGNN        & \textbf{2.98}         & \textbf{6.03}       & \textbf{8.35\%} & 3.19                  & \textbf{6.44}         & 8.69\%                  & 3.26                  & 6.59                  & 9.07\%                   \\
                    & AGCRN        & 3.19                  & 6.47                & 8.81\%          & 3.24                  & 6.60                  & 9.01\%                  & 3.25                  & 6.61                  & 9.19\%                   \\
                    & GTS          & 3.08                  & 6.40                & 8.59\%          & \textbf{3.14}         & 6.52                  & \textbf{7.58\%}         & \textbf{3.12}         & 6.56                  & \textbf{8.61\%}          \\
\cmidrule{2-2}
                    & GRU          & 4.20                  & 7.09                & 11.27\%         & 4.27                  & 7.16                  & 11.42\%                 & 4.45                  & 7.41                  & 11.94\%                  \\
                    & GRU-I        & 4.02                  & 6.83                & 10.89\%         & 4.03                  & 6.88                  & 10.83\%                 & 4.09                  & 6.91                  & 10.88\%                  \\
                    & GRU-D        & 7.50                  & 11.87               & 24.69\%         & 7.45                  & 11.86                 & 24.66\%                 & 7.53                  & 11.91                 & 24.76\%                  \\
                    & LSTM-I       & 4.12                  & 6.89                & 11.04\%         & 4.18                  & 6.98                  & 11.10\%                 & 4.21                  & 7.08                  & 11.26\%                  \\
                    & LSTM-M       & 4.10                  & 6.92                & 10.91\%         & 4.15                  & 6.98                  & 11.03\%                 & 4.26                  & 7.18                  & 11.44\%                  \\
                    & SGMN         & 5.54                  & 10.93               & 13.17\%         & 5.61                  & 10.99                 & 13.35\%                 & 5.81                  & 11.18                 & 13.83\%                  \\
                    & GCN-M (ours) & \uline{3.17}          & \uline{6.33}        & \uline{8.72\%}  & \uline{3.23}          & \uline{6.47}          & \uline{8.99\%}          & \uline{3.26}          & \textbf{\uline{6.35}} & \uline{8.98\%}           \\

\midrule
\multirow{12}{*}{\rotatebox{90}{Long-range missing}} & DCRNN        & 3.46                  & 6.78                & 9.62\%          & 3.54                  & 6.96                  & 9.75\%                  & 3.62                  & 7.02                  & 9.89\%                   \\
                    & STGCN        & 3.71                  & 7.20                & 9.91\%          & 3.76                  & 7.39                  & 10.42\%                 & 3.88                  & 7.66                  & 10.67\%                  \\
                    & GraphWaveNet & 3.43                  & 6.64                & 9.07\%          & 3.57                  & 6.92                  & 9.62\%                  & 3.61                  & 7.03                  & 10.71\%                  \\
                    & MTGNN        & 3.19                  & \textbf{6.32}       & \textbf{8.48\%} & 3.39                  & 6.85                  & 9.21\%                  & 3.50                  & 6.95                  & 9.74\%                   \\
                    & AGCRN        & 3.31                  & 6.54                & 8.94\%          & 3.33                  & 6.68                  & 8.98\%                  & 3.33                  & 6.78                  & 9.45\%                   \\
                    & GTS          & 3.25                  & 6.61                & 8.93\%          & 3.29                  & 6.74                  & 8.85\%                  & 3.46                  & 6.86                  & 9.37\%                   \\
\cmidrule{2-2}
                    & GRU          & 4.37                  & 7.28                & 12.54\%         & 4.47                  & 7.44                  & 11.72\%                 & 4.76                  & 7.81                  & 12.71\%                  \\
                    & GRU-I        & 4.20                  & 6.91                & 11.78\%         & 4.09                  & 6.97                  & 15.42\%                 & 4.21                  & 7.03                  & 11.43\%                  \\
                    & GRU-D        & 7.59                  & 11.94               & 24.72\%         & 7.82                  & 12.46                 & 25.67\%                 & 7.96                  & 12.45                 & 26.12\%                  \\
                    & LSTM-I       & 4.20                  & 7.09                & 11.08\%         & 4.25                  & 7.12                  & 11.32\%                 & 4.36                  & 7.32                  & 11.56\%                  \\
                    & LSTM-M       & 4.53                  & 7.47                & 11.88\%         & 5.21                  & 8.84                  & 15.34\%                 & 6.08                  & 10.02                 & 18.13\%                  \\
                    & SGMN         & 9.47                  & 14.30               & 20.72\%         & 11.49                 & 16.01                 & 24.55\%                 & 13.97                 & 18.24                 & 29.10\%                  \\
                    & GCN-M (ours) & \textbf{\uline{3.18}} & \uline{6.39}        & \uline{8.71\%}  & \textbf{\uline{3.23}} & \textbf{\uline{6.56}} & \textbf{\uline{8.78\%}} & \textbf{\uline{3.27}} & \textbf{\uline{6.68}} & \textbf{\uline{9.12\%}}  \\

\midrule
\multirow{12}{*}{\rotatebox{90}{Mix-range missing}}   & DCRNN        & 3.33                  & 6.69                & 9.53\%          & 3.47                  & 6.85                  & 9.64\%                  & 3.56                  & 6.95                  & 9.78\%                   \\
                    & STGCN        & 3.56                  & 7.12                & 9.81\%          & 3.64                  & 7.28                  & 10.33\%                 & 3.73                  & 7.51                  & 10.62\%                  \\
                    & GraphWaveNet & 3.28                  & 6.51                & 9.02\%          & 3.43                  & 6.78                  & 9.52\%                  & 3.51                  & 6.94                  & 9.62\%                   \\
                    & MTGNN        & \textbf{3.04}         & \textbf{6.18}       & \textbf{7.84\%} & 3.14                  & 6.72                  & 9.07\%                  & 3.44                  & 6.82                  & 9.12\%                   \\
                    & AGCRN        & 3.19                  & 6.49                & 8.77\%          & 3.21                  & 6.56                  & 8.95\%                  & 3.26                  & 6.65                  & 8.98\%                   \\
                    & GTS          & 3.12                  & 6.51                & 8.61\%          & 3.22                  & 6.61                  & 8.84\%                  & 3.34                  & 6.72                  & 8.86\%                   \\
\cmidrule{2-2}
                    & GRU          & 4.30                  & 7.14                & 11.47\%         & 4.35                  & 7.31                  & 11.68\%                 & 4.65                  & 7.72                  & 12.58\%                  \\
                    & GRU-I        & 4.05                  & 6.83                & 10.94\%         & 4.01                  & 6.86                  & 10.83\%                 & 4.11                  & 6.97                  & 10.98\%                  \\
                    & GRU-D        & 7.53                  & 11.89               & 24.74\%         & 7.71                  & 12.32                 & 25.43\%                 & 7.89                  & 12.34                 & 25.63\%                  \\
                    & LSTM-I       & 4.15                  & 6.94                & 11.06\%         & 4.19                  & 7.01                  & 11.18\%                 & 4.30                  & 7.18                  & 11.35\%                  \\
                    & LSTM-M       & 4.40                  & 7.38                & 11.91\%         & 5.14                  & 8.77                  & 14.92\%                 & 6.02                  & 9.92                  & 17.88\%                  \\
                    & SGMN         & 9.33                  & 14.16               & 20.47\%         & 11.42                 & 15.87                 & 24.40\%                 & 13.84                 & 18.13                 & 28.97\%                  \\
                    & GCN-M (ours) & \uline{3.08}          & \uline{6.34}        & \uline{8.59\%}  & \textbf{\uline{3.12}} & \textbf{\uline{6.42}} & \textbf{\uline{8.71\%}} & \textbf{\uline{3.23}} & \textbf{\uline{6.50}} & \textbf{\uline{8.76\%}} \\

\bottomrule
\end{tabular}
}
\end{table}

In Table \ref{BAY_combine_miss} and \ref{LA_combine_miss}, we show the performance comparison on the PEMS-BAY and METR-LA datasets under various missing value scenarios. We highlight the best results among the one-step processing models (underlined values) and all the models (bold values). Globally, GCN-M shows the best performance under all the settings when compared with other one-step processing models. The graph-based model SGMN \citep{cui2020graph} performs much worse than other one-step processing models under long-range and mix-range missing settings, indicating that it applies only to simple missing scenarios, i.e., short-range random missing.
GCN-M does not always show superiority compared with the first group of works, especially in the short-range missing scenario, where MTGNN and GTS usually show good performances. MTGNN typically performs better than GCN-M when the missing rate is low (10\%), except under the mix-range missing scenario of PEMS-BAY. We can draw a conclusion from this observation: a robust Spatio-temporal forecasting model can offset the impact of the missing values to some extent, as it allows exploring the information thoroughly from the observed measures. GCN-M becomes the best forecasting model when the missing rate gets higher, as the missing values become a more critical factor that impacts the forecasting model than Spatio-temporal pattern modeling.

Compared to the short-range missing scenario, GCN-M shows a more robust performance under long-range and mix-range missing scenarios, where the recent temporal values and the nearby nodes' values are not always observed. The multi-scale memory block in GCN-M allows enriching the traffic embedding at each timestamp, thus making the model robust in the two complex scenarios. The memory block searches for the periodic global patterns from historical data and the valuable local features from nearby nodes or recent observations at each timestamp. When nearby node values are unobserved, GCN-M favors more recent observations and vice versa. As the zero values usually show periodicity while missing values show contingency \citep{PEMS_manual}, the memory module with the periodic historical patterns can distinguish the inherent zero values from the missing values. The current node readings combined with the historical patterns will eliminate the effect of missing values but conserve that of zero values.

\begin{figure*}[htbp]

\centering
  \subfloat[][\centering MAE]{
  \hspace{-1cm}
        \includegraphics[width=0.55\linewidth]{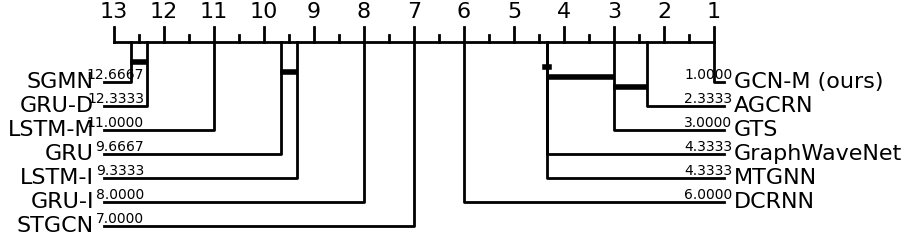}\label{fig_complex_missing_CD_diag:mae}}
  \subfloat[][\centering RMSE
  \hspace{-1cm}
  \label{fig_complex_missing_CD_diag:rmse}]{%
        \includegraphics[width=0.55\linewidth]{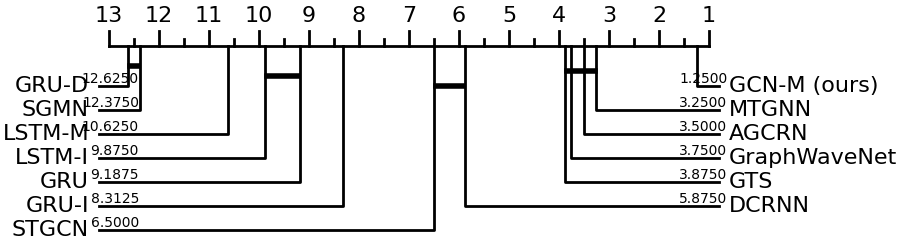}}
        \\
  \subfloat[][\centering MAPE \label{fig_complex_missing_CD_diag:mape}]{%
        \includegraphics[width=0.55\linewidth]{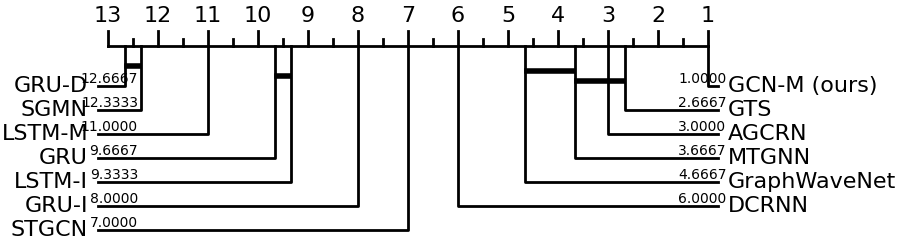}}
    \hfill
  
  \caption{Critical difference diagrams showing rankings by various evaluation metrics and the statistical difference comparison of 13 forecasting models on PEMS-BAY and METR-LA datasets with various missing rates under mix-range missing scenario.}
  \label{fig_complex_missing_CD_diag}
  \vspace{-1.5em}
\end{figure*}

\textcolor{black}{In Figure \ref{fig_complex_missing_CD_diag}, we show the critical difference diagrams \citep{demsar2006statistical} on PEMS-BAY and METR-LA datasets under the mix-range missing scenario. A thick horizontal line shows a group of models that are not significantly different in terms of different evaluation metrics. We can also observe that GCN-M is significantly different from other model groups on all the evaluation metrics, which validates the model's performance for processing missing values under complex scenarios.}

\begin{figure*}
\vspace{-2em}
\centering 
\subfloat[Missing Rate =10\% (L)]{
    \includegraphics[width=0.5\linewidth]{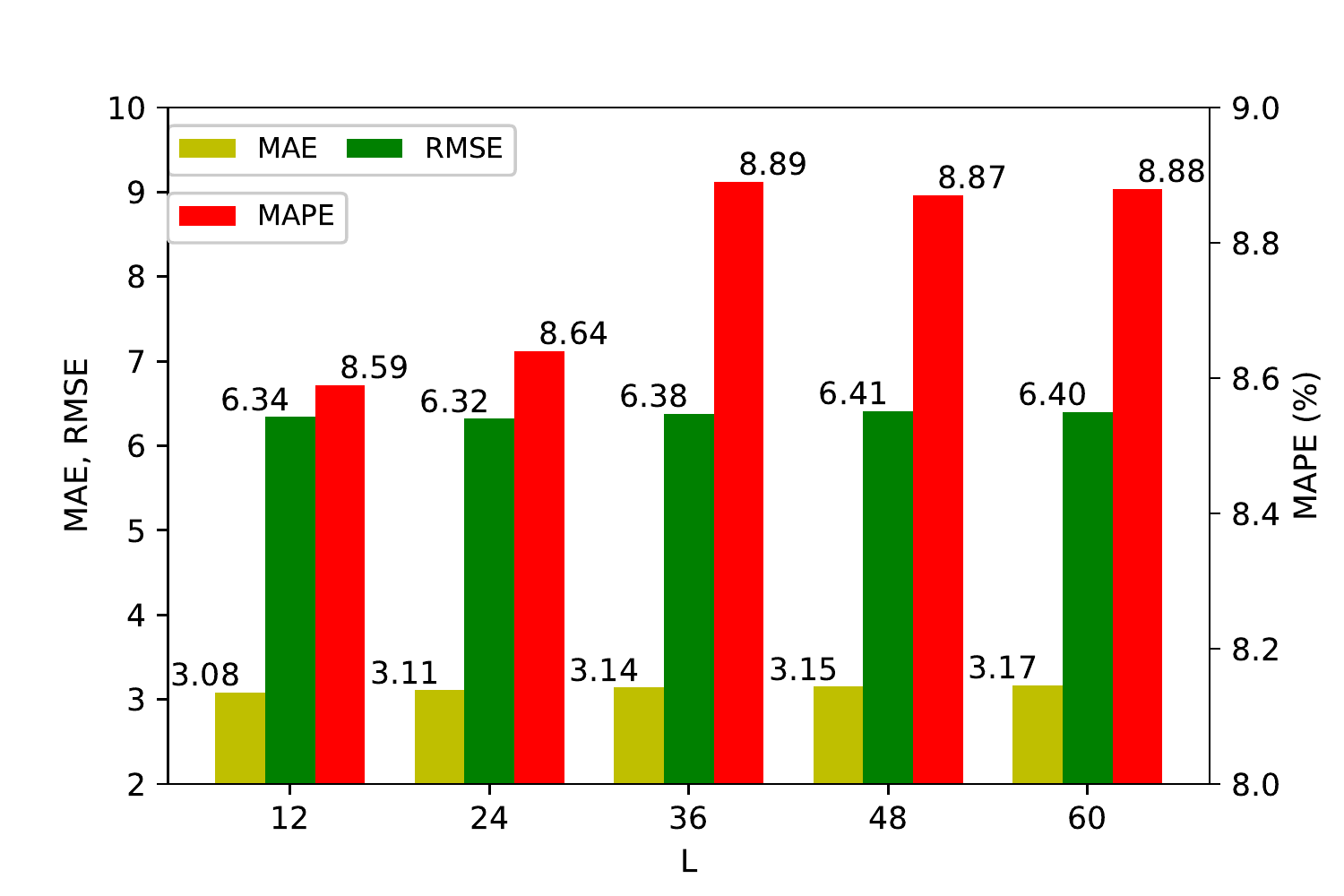}

}
\centering
\subfloat[Missing Rate =20\% (L)]{
    \includegraphics[width=0.5\linewidth]{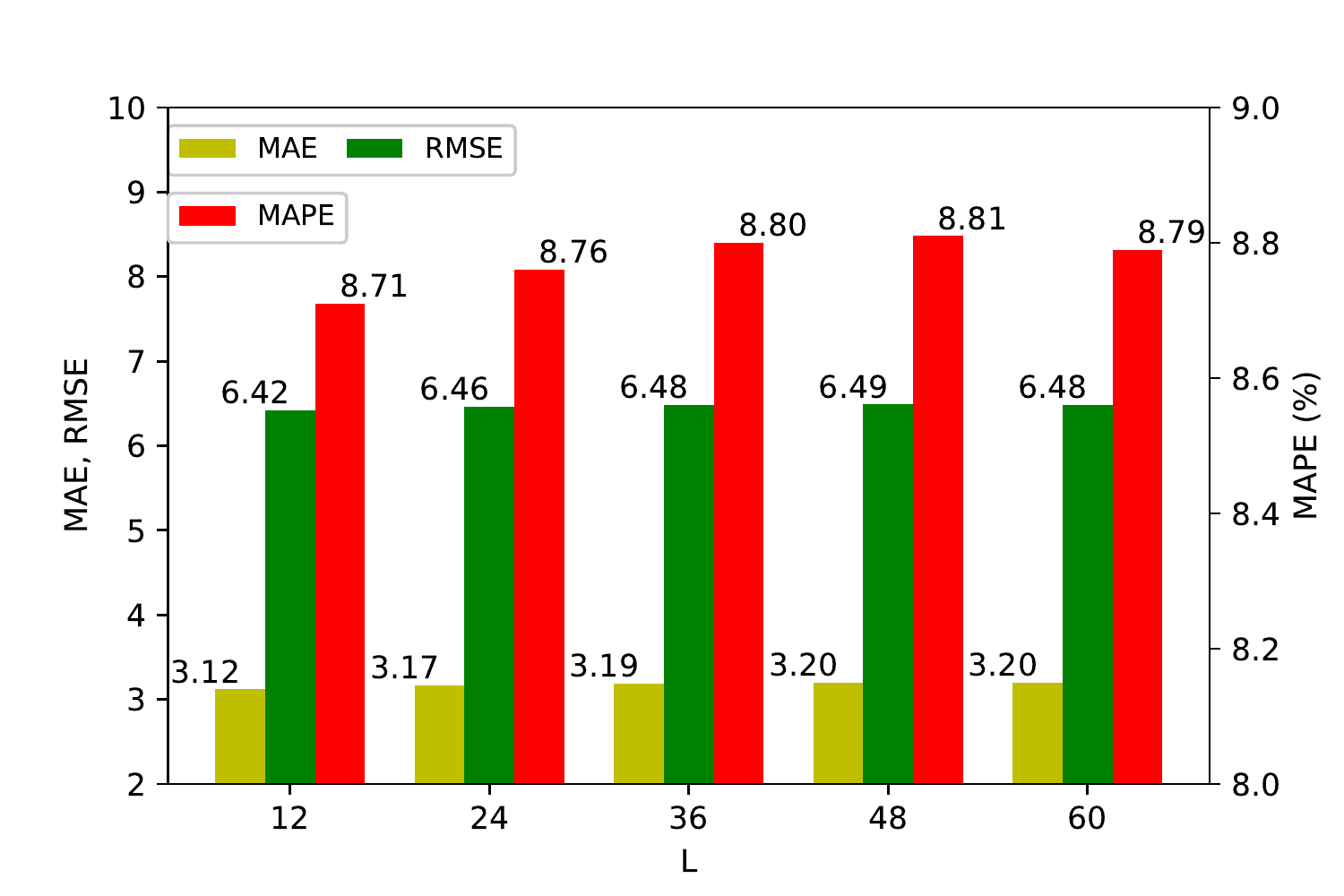}

}
\centering 
\\

\subfloat[Missing Rate =40\% (L)]{
    \includegraphics[width=0.5\linewidth]{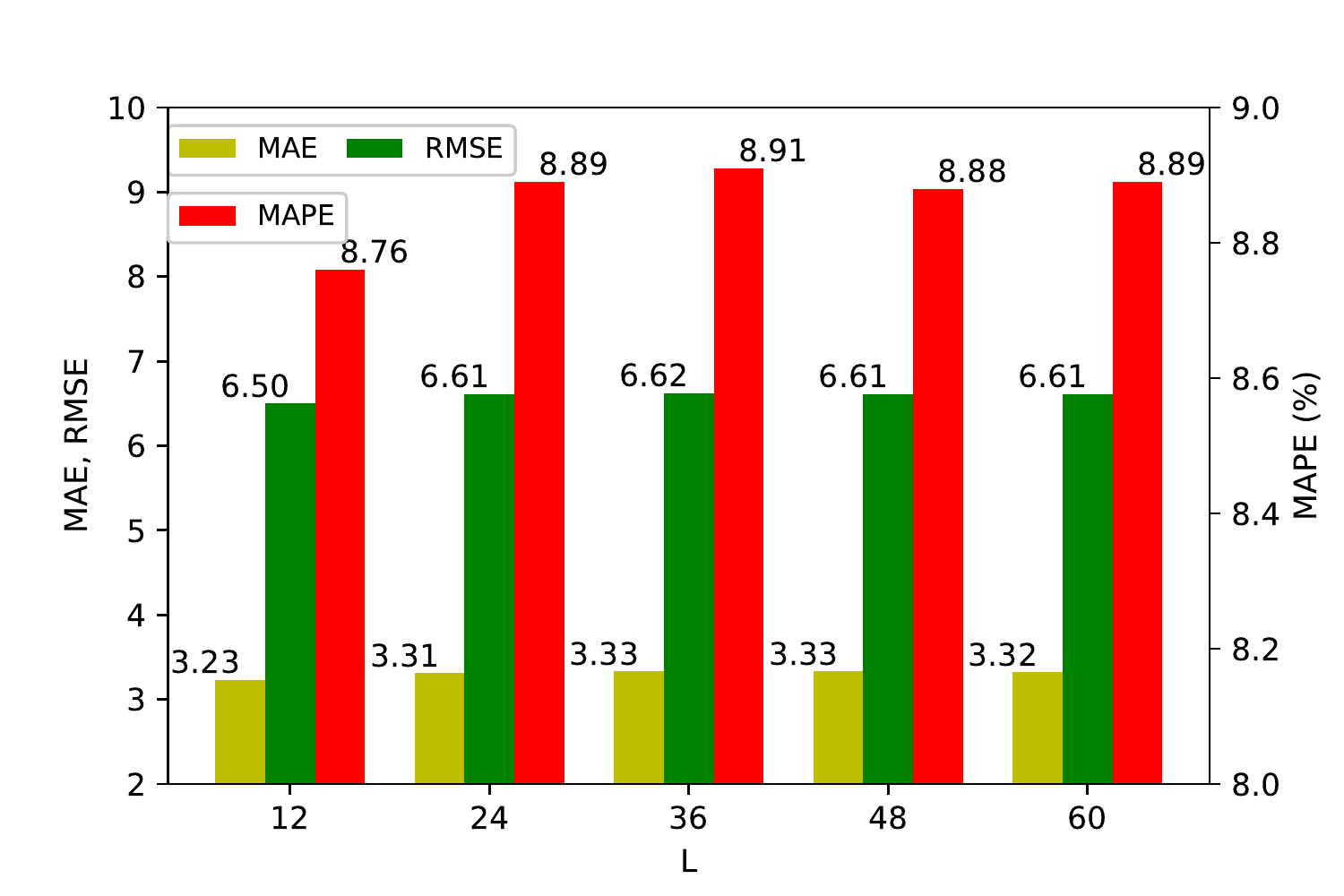}

}
\centering 
\subfloat[Missing Rate =10\% (S)]{
    \includegraphics[width=0.5\linewidth]{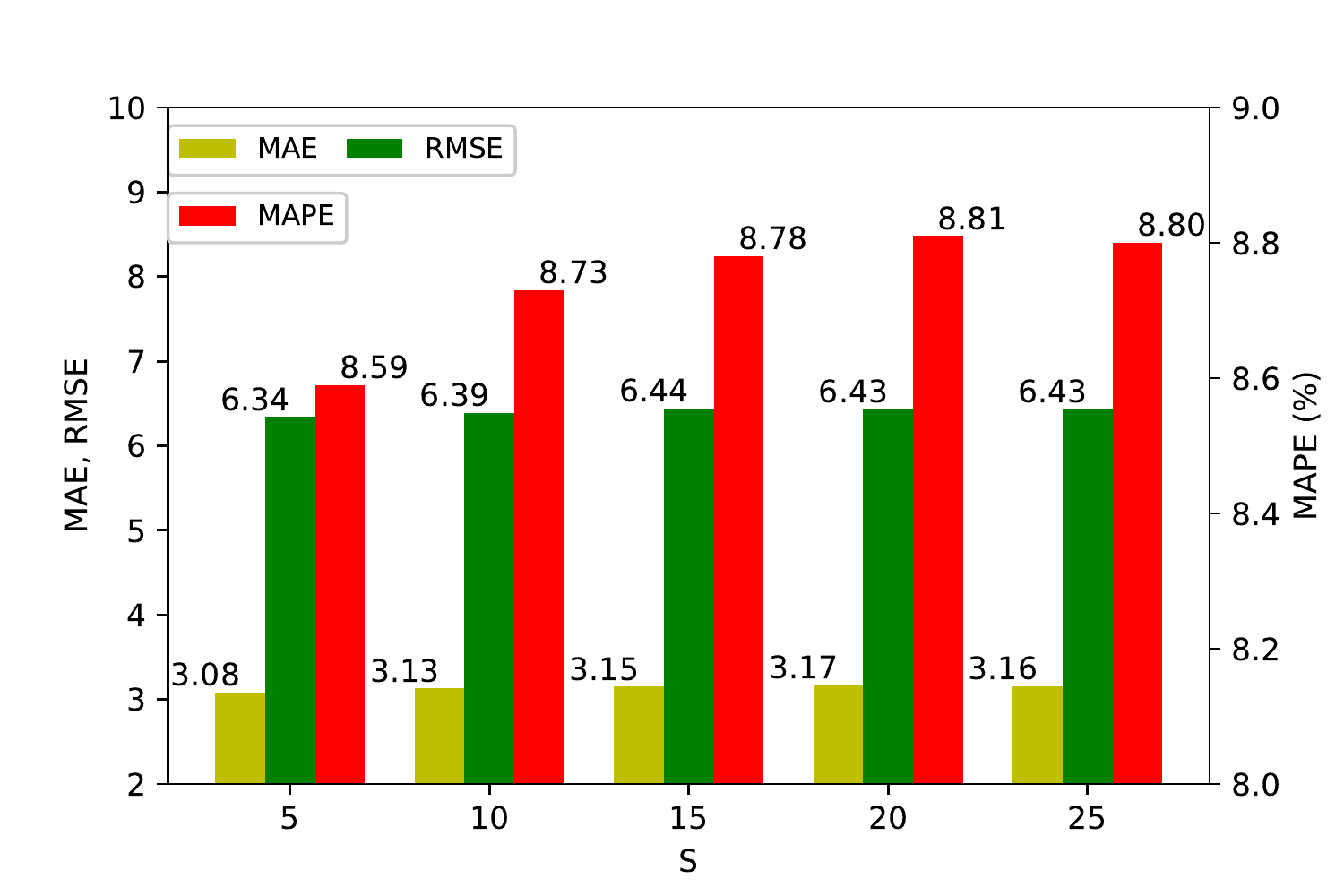}

}
\\

\centering 
\subfloat[Missing Rate =20\% (S)]{
    \includegraphics[width=0.5\linewidth]{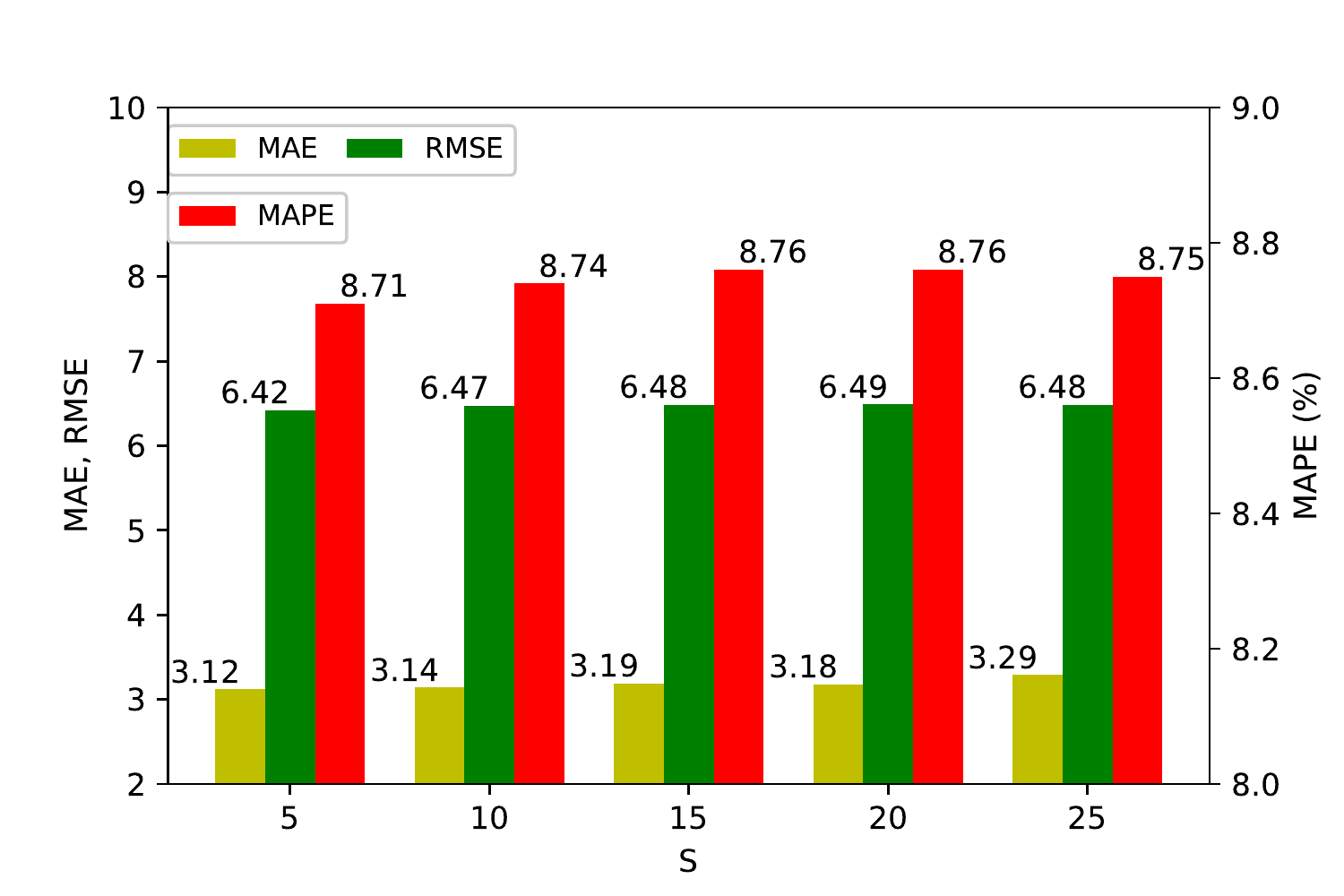}

}
\centering 
\subfloat[Missing Rate =40\% (S)]{
    \includegraphics[width=0.5\linewidth]{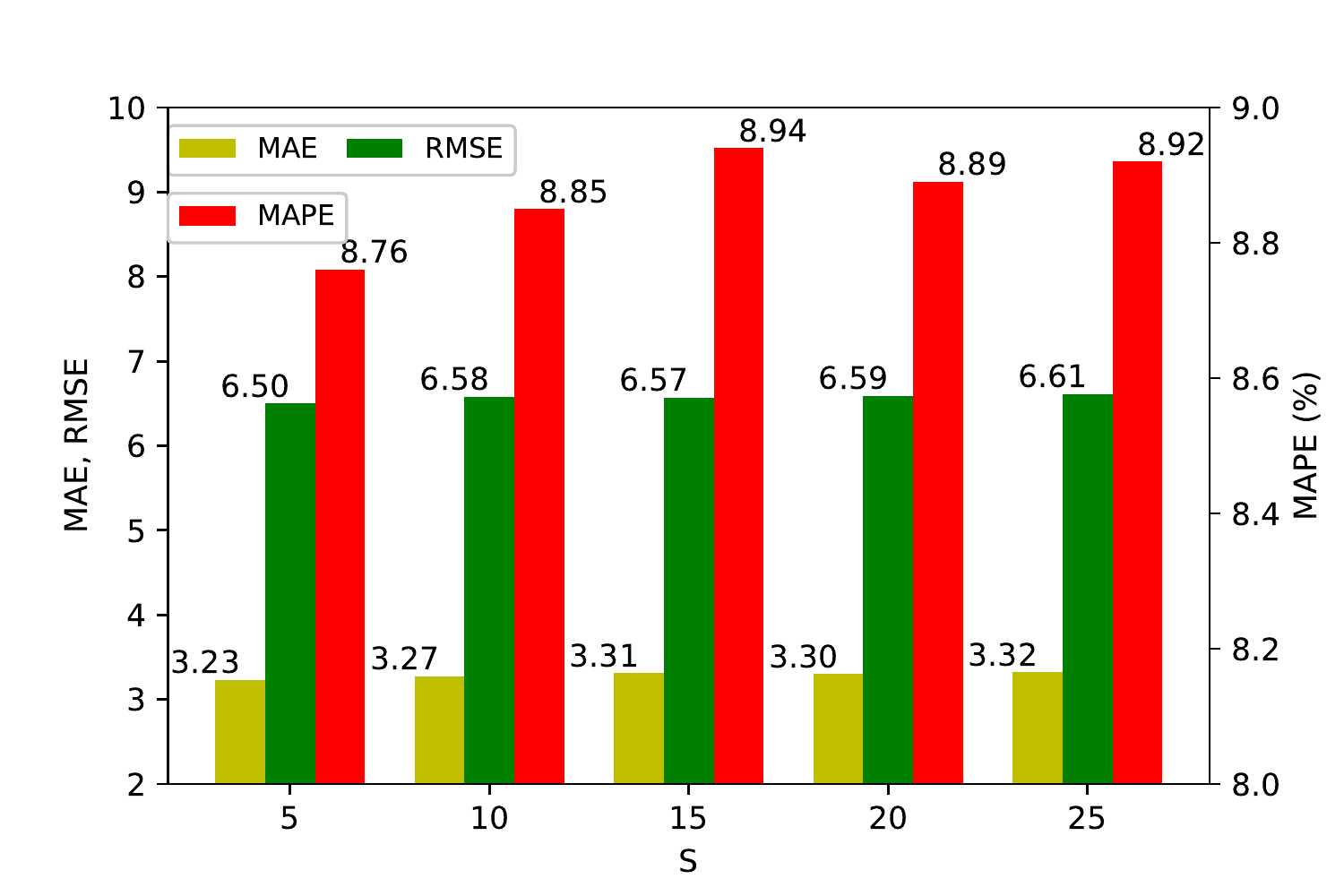}

}
\hfill
\caption{Parameters effects: we report the model errors of GCN-M on METR-LA dataset considering \textit{a,b,c)} $L$ observed samples before current timestamp for constructing the \textit{Empirical Temporal Mean} in equation \ref{temp_mean_equation}; \textit{d,e,f)} $S$ observed samples nearby current node for constructing the \textit{Empirical Spatial Mean} in equation \ref{spatial_mean_equation}.}
\label{parameter_effects}
\vspace{-1em}
\end{figure*}

In Figure \ref{parameter_effects}, we show the effects of the memory module's parameters $L$ and $S$ on the model's performance. The two parameters represent the searching range of the local temporal and spatial features, respectively. We report the model's evaluation errors with various missing rates. From the results in Figure \ref{parameter_effects}, we observe that when the searching range becomes more extensive, the model's performance decreases more. This can be explained by the mean value of a larger space and the less recent observations will lead to a weaker information dependency with the current timestamp, thus affecting the information enrichment of the traffic embedding. In real-life datasets, we can set the parameters from a small value, such as considering local features during the last one hour (L=12) with five nearest sensor nodes (S=5).

\subsection{RQ 3: Dynamic Graph Modeling}
In the dynamic traffic system, the spatial dependency can be considered as a dynamic system status, which evolves over time \citep{han2021dynamic}. The traffic observations at each timestamp are always adopted to characterize the dynamic traffic status and help learn the dynamic graphs \citep{li2021dynamic}. However, due to the missed observations, the traffic status at certain timestamps can not be characterized, thus affecting the dynamic graph learning process. 

This issue can be handled by the enriched traffic embeddings proposed in GCN-M. It allows considering the local static features and global historical patterns, which avoids the deviation introduced by the missing values and helps learn the dynamic graphs. To validate the performance of the learned dynamic graphs, we designed the following variants of our GCN-M model:
\begin{itemize}
    \item GCN-M-obs: instead of using the enriched traffic embeddings, the raw traffic observations \citep{li2021dynamic} are adopted to construct dynamic graphs.
    \item GCN-M-adp: instead of learning dynamic graphs and applying dynamic convolution, an adaptive static graph \citep{wu2019graph} is learned to do the graph convolution.
    \item GCN-M-pre: instead of learning graphs from the traffic embedding or observations, the predefined graphs \citep{li2018diffusion} calculated with the directed distances between traffic nodes are adopted for doing graph convolution.
    \item GCN-M-com: combine both predefined and learned static graphs \citep{wu2019graph} to do the graph convolution. 
\end{itemize}

We show in Table \ref{dynamic_graph_results} the performance comparison on various model variants of the spatial graph modeling. We report the model errors on multiple horizons. We consider the complex scenario of mix-range missing values with a missing rate of 40\% on both PEMS-BAY and METR-LA datasets. The results in Table \ref{dynamic_graph_results} suggest that the dynamic graphs learned from the enriched traffic embeddings perform the best when compared to other variants. In contrast, the model obtains the worst performance when learning the dynamic graphs from the raw observations, which is mainly due to the missing values hindering the graph learning process in inferring the dynamic traffic status. GCN-M-obs performs even worse than GCN-M-adp in which the static graph is learned from the entire observations, eliminating the effect from local missing values. 

\begin{table}[htbp]
\centering
\caption{Performance comparison on graph module in mix-range missing value scenario with missing rate = 40\%. \textcolor{black}{The results show the forecasting errors at each forecasting step (i.e., horizon).}}
\label{dynamic_graph_results}
\scalebox{0.8}{
\begin{tabular}{lllllllllll}
\toprule
         &              & \multicolumn{3}{c}{Horizon=3 (15 mins)}         & \multicolumn{3}{c}{Horizon=6 (30 mins)}         & \multicolumn{3}{c}{Horizon=12 (60 mins)}  \\
         & Models       & MAE           & RMSE          & MAPE            & MAE           & RMSE          & MAPE            & MAE           & RMSE          & MAPE             \\
\midrule
\multirow{5}{*}{\rotatebox{90}{PEMS-BAY}}
 & GCN-M-obs & 1.63          & 3.48          & 3.53\%          & 1.93          & 4.16          & 4.31\%          & 2.25          & 4.81          & 5.10\%           \\
         & GCN-M-adp      & 1.53          & 3.11          & 3.14\%          & 1.82          & 3.93          & 4.03\%          & 2.14          & 4.72          & 4.92\%           \\
         & GCN-M-pre     & 1.61          & 3.27          & 3.21\%          & 1.87          & 4.05          & 4.11\%          & 2.18          & 4.74          & 5.03\%           \\
         & GCN-M-com    & 1.54          & 3.13          & 3.11\%          & 1.79          & 3.92          & 3.97\%          & 2.11          & 4.62          & 3.91\%           \\
         & \textbf{GCN-M}        & \textbf{1.45} & \textbf{3.03} & \textbf{3.09\%} & \textbf{1.70} & \textbf{3.81} & \textbf{3.89\%} & \textbf{2.06} & \textbf{4.64} & \textbf{4.86\%}  \\
\midrule
\multirow{5}{*}{\rotatebox{90}{METR-LA}}
  & GCN-M-obs & 2.97          & 5.68          & 7.71\%          & 3.31          & 6.57          & 8.78\%          & 3.71          & 7.54          & 10.07\%          \\
         & GCN-M-adp      & 2.84          & 5.51          & 7.44\%          & 3.19          & 6.41          & 8.59\%          & 3.68          & 7.34          & 9.96\%           \\
         & GCN-M-pre     & 2.92          & 5.56          & 7.64\%          & 3.23          & 6.42          & 8.63\%          & 3.72          & 7.42          & 10.04\%          \\
         & GCN-M-com    & 2.84          & 5.52          & 7.45\%          & 3.17          & 6.4           & 8.63\%          & 3.68          & 7.41          & 9.97\%           \\
         & \textbf{GCN-M}        & \textbf{2.82} & \textbf{5.47} & \textbf{7.42\%} & \textbf{3.16} & \textbf{6.38} & \textbf{8.55\%} & \textbf{3.58} & \textbf{7.31} & \textbf{9.92\%} \\
         \bottomrule
\end{tabular}
}
\end{table}

\color{black}\subsection{RQ 4: One-step VS two-step processing}
In this section, we show the performance comparison when adopting different missing-value processing strategies for traffic forecasting. As a one-step processing model, GCN-M jointly models the Spatio-temporal patterns and missing values for traffic forecasting. The two-step processing models handle the missing values in a preprocessing step, then apply a forecasting model to the completed data.

To compare the processing strategies fairly, we use GCN-M as a base model to test the two-step processing approaches. A preprocessing step is adopted to replace the GCN-M memory module designed to handle missing values.

We consider the following imputation methods to fill in missing values and apply GCN-M to the completed data.
\begin{itemize}
        \item MEAN \citep{garcia2010pattern}: This approach replaces missing values with the mean of observed measures based on the respective feature of the input sequence.
        \item KNN \citep{batista2002study}: This method replaces missing values with the mean of k-nearest temporal neighbors. We linearly interpolate the missing values with k = 2, considering the previous and next non-empty values.
        \item MICE \citep{van2011mice}: MICE is the multiple imputation method that fills the missing values from the conditional distributions by Markov chain Monte Carlo (MCMC) techniques.
\end{itemize}

We show in Table \ref{one_two_step_processing} the performance comparison between one-step processing (i.e., GCN-M) and two-step processing (i.e., GCN-M variants) models. We conducted the experiments under the complex scenario of missing values (i.e., mix-range missing) with missing rates varying from 10\% to 40\%. Table \ref{one_two_step_processing} shows that the preprocessing step with different imputation techniques leads to worse performance than the one-step setting. Even though authors in \cite{shleifer2019incrementally} justified that the imputation techniques in the preprocessing step improve the model's performance under simple scenarios of missing values (e.g., random short-range missing), they are not applicable for complex scenarios of missing values with the results obtained from Table \ref{one_two_step_processing}. On the one hand, the complex missing values (e.g., on short \& long ranges, partial \& entire variables) challenges the imputation tasks in considering local and global Spatio-temporal patterns; on the other hand, the patterns of missing values modeled in the preprocessing step are totally isolated from forecasting models, which may not be valuable for the forecasting tasks.

\begin{table}[htbp]
\centering
\caption{Performance comparison between one-step and two-step processing models under mix-range missing value scenario with various missing rates. The results show a one-hour (12-step) average of the forecasting errors. }
\label{one_two_step_processing}
\scalebox{0.8}{
\renewcommand{\arraystretch}{1.12} % Default value: 1
\begin{tabular}{lllllllllll}
\toprule

                  &                & \multicolumn{3}{c}{Missing Rate = 10\%}         & \multicolumn{3}{c}{Missing Rate = 20\%}         & \multicolumn{3}{c}{Missing Rate = 40\%}          \\
                  & Models         & MAE           & RMSE          & MAPE            & MAE           & RMSE          & MAPE            & MAE           & RMSE          & MAPE             \\ 
\midrule
\multirow{4}{*}{\rotatebox{90}{PEMS-BAY}} & MEAN-GCN-M     & 1.72          & 3.88          & 3.90\%          & 1.82          & 4.01          & 4.08\%          & 1.97          & 4.35          & 4.45\%           \\
                  & KNN-GCN-M      & 1.74          & 3.86          & 3.93\%          & 1.84          & 3.99          & 4.06\%          & 2.06          & 4.39          & 4.55\%           \\
                  & MICE-GCN-M     & 1.75          & 3.91          & 3.98\%          & 1.82          & 3.94          & 4.04\%          & 1.95          & 4.28          & 4.39\%           \\
                  & \textbf{GCN-M} & \textbf{1.65} & \textbf{3.67} & \textbf{3.69\%} & \textbf{1.66} & \textbf{3.72} & \textbf{3.62\%} & \textbf{1.69} & \textbf{3.79} & \textbf{3.83\%}  \\
\midrule
\multirow{4}{*}{\rotatebox{90}{METR-LA}} & MEAN-GCN-M     & 3.24          & 6.61          & 9.42\%          & 3.58          & 7.12          & 9.83\%          & 3.76          & 7.46          & 10.21\%          \\
                  & KNN-GCN-M      & 3.19          & 6.63          & 9.48\%          & 3.62          & 7.26          & 9.91\%          & 3.81          & 7.52          & 10.38\%          \\
                  & MICE-GCN-M     & 3.14          & 6.52          & 8.92\%          & 3.52          & 6.87          & 9.68\%          & 3.72          & 7.54          & 10.12\%          \\
                  & \textbf{GCN-M} & \textbf{3.08} & \textbf{6.34} & \textbf{8.59\%} & \textbf{3.12} & \textbf{6.42} & \textbf{8.71\%} & \textbf{3.23} & \textbf{6.50} & \textbf{8.76\%}  \\
         \bottomrule
\end{tabular}
}
\end{table}

\color{black}
\subsection{Discussions}
Our approach has several advantages. First, starting from the real-world data, GCN-M considered the complex scenarios of missing values in traffic data. Different from the previous work \citep{che2018recurrent,cui2020graph,tian2018lstm} which consider the missing value from a part of the real-life scenarios: either under the short-range or long-range missing settings, under partial or entire networking missing settings. GCN-M considers the complex mix-missing value context covering various real-life scenarios for missing values. 

Second, GCN-M is capable of handling such complex missing value scenarios with a multi-scale memory module. This combines local Spatio-temporal features (short-range missing, partial and entire network missing) and global historical patterns (long-range missing) to generate the enriched traffic embeddings. The embeddings allow distinguishing the inherent zero values from the missing values. In this way, GCN-M jointly models the Spatio-temporal patterns and missing values in one-step processing, which generally allows a better model performance than two-step processing \citep{cui2020graph}. 

Third, GCN-M allows generating reliable dynamic graphs from the enriched traffic embeddings, which opens a path for learning robust dynamic graphs under missing value settings. Moreover, the generated dynamic graphs can cooperate with various advanced graph convolution modules \citep{wu2020connecting} to improve the model's performance further. 

Last but not least, even though GCN-M is designed for traffic forecasting, it is applicable to wider application domains sharing similar Spatio-temporal characteristics and missing-value scenarios, such as crowd flow forecasting \citep{xie2020urban}, weather and air pollution forecasting \citep{han2021joint, el2022learning, abboud2021micro}, etc. The Spatio-temporal patterns in those data and the missing values caused by the sensor issues or control center errors form similar research problems to this paper. 

However, GCN-M does have a limitation in terms of computational efficiency. Table \ref{modelefficiency} shows the per epoch training time comparison on the full datasets between GCN-M and the baseline models. The one-step processing baseline models are much more efficient than other models. This is basically because of their simple structure without integrating the costly graph convolution modules. GCN-M still performs better than DCRNN, but worse than other forecasting models. This is mainly caused by two factors: 1) generating the enriched traffic embeddings requires a huge computation cost on the attention score's calculation in the memory module; 2) generating the dynamic graphs for graph convolution requires learning a large number of parameters, thus increasing computation cost. Possible solutions might be to reduce the time complexity for calculating the attention with \textit{ProbSparse Attention} proposed in \cite{zhou2021informer}, and to apply more efficient dynamic graph convolution such as graph tensor decomposition \citep{han2021dynamic} and \textcolor{black}{node sampling when generating the graphs \citep{wu2020connecting}}.

\begin{table}[htbp]
\centering
\caption{Model efficiency: training time per epoch (s)}
\label{modelefficiency}
\scalebox{0.8}{
\begin{tabular}{cccccc}
\toprule
Models        & PEMS-BAY & METR-LA & Models  & PEMS-BAY & METR-LA  \\
\midrule
DCRNN       & 468.22   & 178.23  & GRU     & 3.65     & 2.45     \\
STGCN         & 55.32    & 27.70   & GRU-I   & 4.22     & 3.67     \\
GraphWaveNet  & 118.77   & 48.16   & GRU-D   & 7.82     & 5.43     \\
MTGNN         & 86.20    & 38.70   & LSTM-I  & 4.32     & 4.64     \\
AGCRN         & 67.40    & 32.9    & LSTM-M  & 8.12     & 5.76     \\
GTS           & 191.4    & 62.3    & SGMN    & 3.45     & 2.38     \\
GCN-M (ours)  & 241.69   & 118.65  &  -     &  -        &  -  \\
\bottomrule
\end{tabular}
}
\end{table}

\section{Conclusion}
In this paper, we propose GCN-M, a graph convolutional network-based model for handling complex missing values in traffic forecasting. We studied the complex scenario where missing traffic values occur on both short \& long ranges and on partial \& entire transportation networks. The enriched traffic embeddings learned by a Spatio-temporal memory module allow handling the complex missing values and constructing dynamic traffic graphs to improve the model's performance. A joint model optimization is applied to consider missing values and traffic forecasting in one-step processing. We compare GCN-M with the one-step processing models, which are specifically designed for processing incomplete traffic data and the recent advanced traffic forecasting models. The extensive experiments on two benchmark traffic datasets with 12 baselines demonstrate that GCN-M shows a clear advantage under various scenarios of complex missing values, as compared to the advanced traffic forecasting models, while at the same time maintaining comparable performance on complete traffic datasets. These experiments also provide an up-to-date comparison of the traffic forecasting models would it be with or without missing values.
In future work, we will explore the aforementioned optimizations to reduce computational costs. From a longer-term perspective, one can consider noisy data or external events that may impact the predictions.

\section*{Acknowledgements}
This research was supported by DATAIA convergence institute as part of the \textit{Programme d’Investissement d'Avenir}, (ANR-17-CONV-0003) operated by DAVID Lab, University of Versailles Saint-Quentin, and the MASTER project that has received funding from the European Union’s Horizon 2020 research and innovation programme under the Marie-Sklodowska Curie grant agreement N. 777695. The authors would like to thank as well the publication support from the Technology Innovation Institute.

%\bibliographystyle{sn-basic} 
%%\bibliography{reference.bib}% common bib file


\end{document}